\pdfoutput=1

\documentclass[11pt]{article}

\usepackage[final]{acl}

\usepackage{times}
\usepackage{latexsym}
\usepackage{tikz}
\usepackage{makecell}
\usepackage{siunitx}
\usepackage{algorithm}
\usepackage{algpseudocode} %
\usepackage{multirow}

\usepackage{subcaption} %

\usepackage[T1]{fontenc}

\usepackage[utf8]{inputenc}

\usepackage{microtype}
\usepackage{booktabs}
\usepackage{pifont}

\usepackage[table]{xcolor}
\definecolor{myred}{HTML}{B22222}
\definecolor{navyblue}{rgb}{0.0, 0.0, 0.5}
\hypersetup{colorlinks, 
            citecolor=myred}

\newcommand{\Ours}{\textsc{\textbf{RPO}}}

\usepackage{inconsolata}

\usepackage{graphicx}
\usepackage{amsmath}

\usepackage{amsfonts} %
\usepackage{empheq} %
\usepackage{mathrsfs}
\usepackage{enumitem}
\usepackage{tabularx} %
\usepackage{ragged2e} %
\usepackage{threeparttable} %
\usepackage{float}

\usepackage{tcolorbox}
\tcbuselibrary{skins}  
\usepackage{lipsum}

\definecolor{myLightGrayBg}{RGB}{245, 245, 245}  %
\definecolor{myMediumGrayHeaderBg}{RGB}{220, 220, 220} %
\definecolor{myDarkGrayBorder}{RGB}{180, 180, 180}   %

\colorlet{currentDefinitionBodyBg}{myLightGrayBg}
\colorlet{currentDefinitionTitleBg}{myMediumGrayHeaderBg}
\colorlet{currentDefinitionBorder}{myDarkGrayBorder}

\newtcolorbox{mydefinition}[2][]{%
  skin=bicolor,                 %
  colback=currentDefinitionBodyBg,    %
  colframe=currentDefinitionBorder,   %
  colbacktitle=currentDefinitionTitleBg, %
  fonttitle=\bfseries,
  coltitle=black,                 %
  title=#2,
  breakable,
  enhanced,
  arc=2mm,                        %
  boxrule=0.8pt,                  %
  fuzzy shadow={0pt}{-1pt}{-0.5pt}{0.5pt}{black!20!gray!30!white}, %
  #1
}

\usepackage{amsthm}

\newcommand{\xmark}{\ding{55}}
\newcommand{\cmark}{\ding{51}}

\newtcolorbox{takeawaybox}[2][]{
    colback=blue!5!white,  %
    colframe=black,        %
    colbacktitle=black,    %
    coltitle=white,        %
    fonttitle=\bfseries,   %
    title=#2,              %
    arc=2mm,               %
    #1
}

\newcommand{\Autoref}[1]{%
  \begingroup%
  \autoref{#1}%
  \def\algorithmautorefname{Algorithm}%
  \def\chapterautorefname{Chapter}%
  \def\sectionautorefname{Section}%
  \def\subsectionautorefname{Subsection}%
  \endgroup%
}
\newcommand{\Tableautoref}[1]{%
  \begingroup%
  \def\figureautorefname{Table}%
  \endgroup%
}

\title{Efficiently Learning To Reason or Not to Reason:\\Root-token Policy Optimization for Adaptive Thinking}

\author{
  Taehyeon~Kim\textsuperscript{$\spadesuit$}\quad
  Hyunsoo~Lee\textsuperscript{$\spadesuit$}\quad
  Youngsoo~Jang\textsuperscript{\ddag,}$^{\dag}$\textsuperscript{,}\thanks{Work completed while employed at LG AI Research.}\quad
  Moontae~Lee\textsuperscript{$\spadesuit$,\S,}\thanks{Corresponding authors.}\\[6pt]
  \textsuperscript{$\spadesuit$}LG AI Research\quad
  \textsuperscript{\ddag}UNIST\quad
  \textsuperscript{\S}University of Illinois Chicago
}

\begin{document}
\maketitle
\begin{abstract}
Large reasoning models (LRMs) achieve strong performance by externalizing explicit reasoning traces before producing the answer,
yet suffer from \textit{overthinking} challenge that allocates uniformly heavy computation to queries of varying difficulty. 
While proprietary models mitigate this via opaque routing, open-source LRMs still lack an efficient mechanism to internalize adaptive reasoning due to both expensive training cost and limited disclosure of training recipes. 
In response, we introduce \Ours~(\underline{R}oot-token \underline{P}olicy \underline{O}ptimization), a framework that enables LRMs to self-determine when to reason by training only the initial root token (e.g., whether to invoke the \texttt{<think>} tag) via group relative reward and group-wise advantages. 
By focusing on this pivotal branching point, \Ours~drastically reduces training overhead and VRAM usage. Across multiple model families and scales, \Ours~learns difficulty-aware adaptive thinking at just $\sim$2\% of the training compute of prior adaptive-reasoning methods.
\end{abstract}

\section{Introduction}\label{introduction}

The advent of large reasoning models (LRMs) has redefined the benchmark requiring complex reasoning \cite{guo2025deepseek, jaech2024openai}. 
The efficacy of LRMs has been attributed to the additional computation allocated at test time through explicit reasoning paths, typically encapsulated within delimiters such as \texttt{<think>} ... \texttt{</think>}. By exploring logical traces, reflecting on internal errors, and self-verifying intermediate steps, these models have achieved breakthroughs in mathematical and symbolic reasoning tasks. These days, this \textit{think-before-answer} paradigm has emerged as the de facto standard.

\begin{figure}[t]
    \centering
    \includegraphics[width=\linewidth]{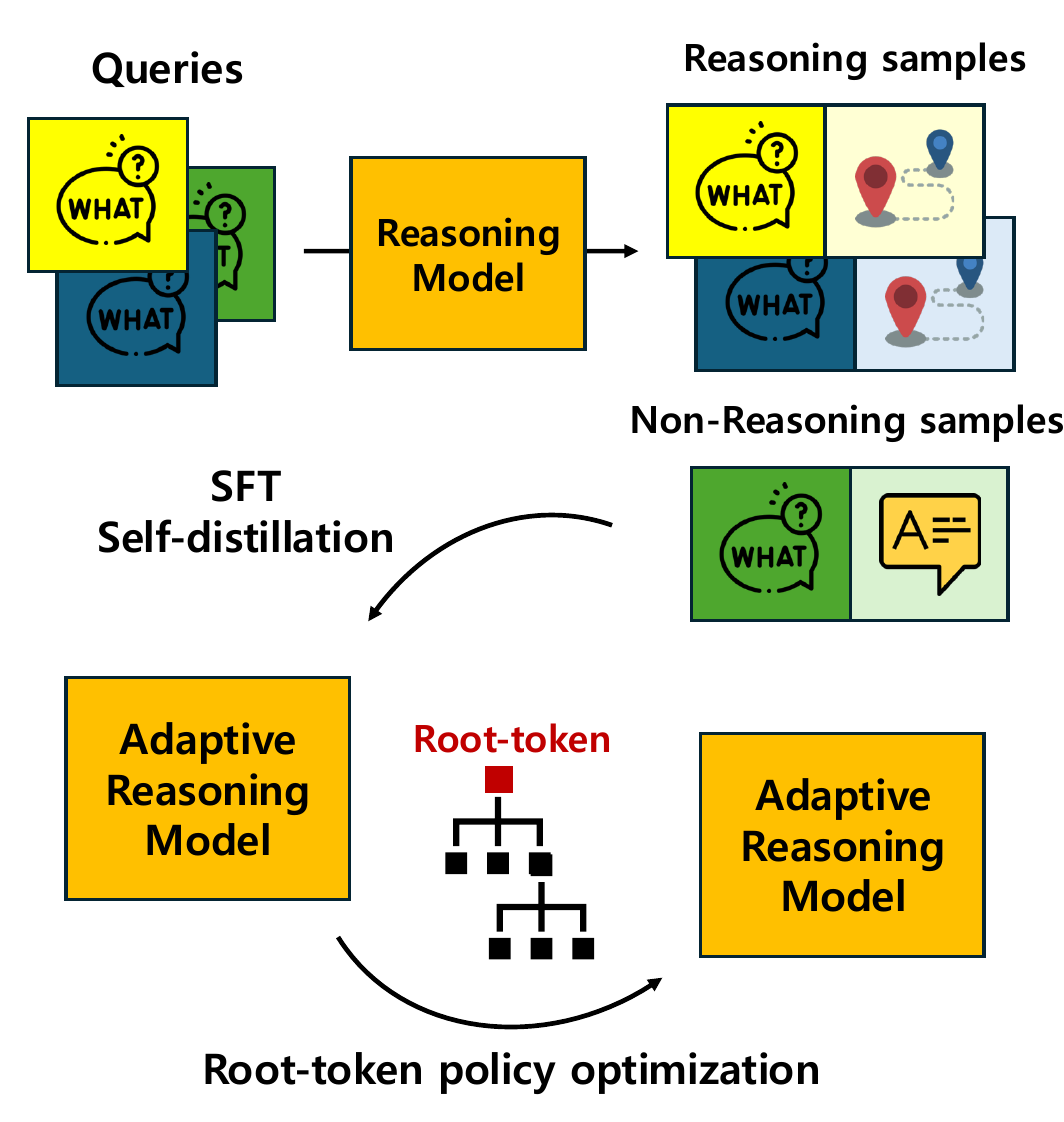}
    \caption{Overview of our training pipeline. We generate paired reasoning samples and non-reasoning samples from input queries through the original reasoning model.
    These samples are self-distilled to manifest adaptive reasoning. Finally, root-token policy optimization (\Ours) is applied to the model, refining the policy by updating only the initial branching token (i.e., root-token) to improve difficulty-aware adaptive reasoning.
    }
    \label{fig:landing}
\end{figure}

\begin{figure*}[t]
    \centering
    \includegraphics[width=0.98\linewidth]{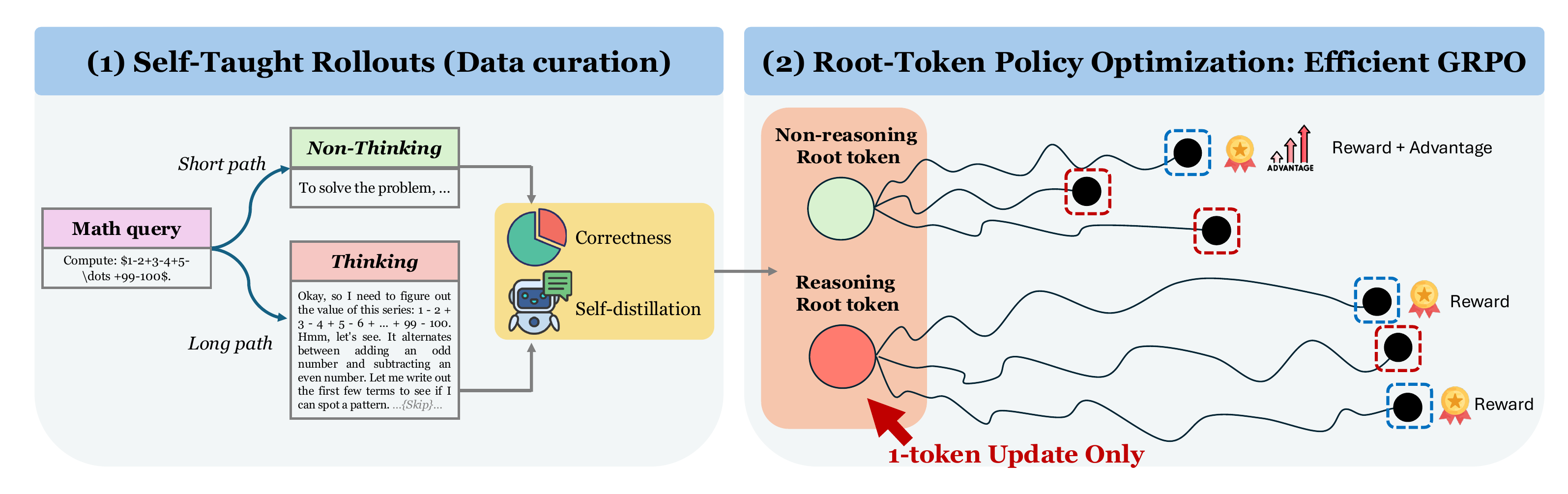}
    \caption{Detailed overview of our training framework. After supervised finetuning (SFT) with self-generated correct samples, \Ours~involves updates only the initial decision token with group normalized rewards and advantages.}
    \label{fig:overview}
\end{figure*}

Despite the potential of LRMs, there remains a significant limitation that general LRMs tend to apply this exhaustive explicit reasoning uniformly across all queries, regardless of their intrinsic complexity.
This results in the \textit{overthinking} problem \cite{feng2025efficient, sui2025stop} where substantial computational resources and latency are wasted on trivial questions. Consequently, a surge of emerging research has been driven into striking a balance between reasoning quality and computational efficiency \cite{Zhang2025AdaptThink, Bao2025Thinkless, wu2025arm, wang2025adareasoner}.
While proprietary services such as GPT5 \cite{openai2025gpt5} and Claude \cite{Claude3_5SonnetSystemCard} rely on black-box routing to mitigate overthinking, a novel model-centric alternative is required to internalize adaptive reasoning via policy optimization with open dataset for the open-source community.

While adaptive reasoning models that learn when to reason have been explored via policy optimization, these studies have faced the following challenges:

\begin{enumerate}[leftmargin=12pt, itemsep=2pt]
    \item Training inefficiency. Optimizing over long context trajectories with policy optimization incurs prohibitive VRAM and temporal costs, hindering accessible and iterative development.

    \item Data curation. There is a lack of standardized methods for \textit{baking} difficulty-aware datasets that effectively balance reasoning traces with direct responses across heterogeneous domains.

    \item Initialization gap. Prior works train from scratch or adapt base models, often avoiding adapting the original LRM due to its training sensitivity.

\end{enumerate}

To alleviate these challenges, we propose \Ours~(\underline{R}oot-token \underline{P}olicy \underline{O}ptimization), an efficient training method that enables adaptive reasoning by learning only the initial root token (e.g., the decision to open or close the \texttt{<think>} tag). Our method involves two stages (\Autoref{fig:landing}): (1) offline synthesis which generates paired reasoning and non-reasoning responses via forced chat templates, and (2) \Ours, which restricts gradient updates to the pivotal branching decision. 
Our design is complementary to recent token-selective RL that updates only a small subset of decision tokens (e.g., high-entropy forking tokens) rather than the full trajectory~\cite{wang2025beyond8020}; \Ours~pushes this idea to the extreme by optimizing only the root routing token (\Autoref{fig:overview}). 
We demonstrate that \Ours~drastically enhances memory and compute efficiency during training while preserving the model's original capability. 

Our contributions are as follows:

\begin{itemize}[leftmargin=12pt, itemsep=2pt]
    \item We propose \Ours, a novel framework, termed as root-token optimization to enable adaptive reasoning by optimizing only the initial root-token whether to reason, significantly reducing training overhead as well as efficient difficulty-aware routing (\S\ref{sec:method}).

    \item Our empirical results demonstrate \Ours~enables LRMs learning when to reason over diverse LRM families and scales. We also provide an analysis on calibration for an explicit reasoning knob (e.g., target Think\% or routing threshold for root-token probability) (\S\ref{sec:exp}, \S\ref{sec:Analysis}).

    \item We demonstrate that difficulty-aware policies in mathematical domains elicit adaptive reasoning in broader scientific and code domain (\S\ref{sec:Analysis}).
\end{itemize}

\section{Related work}\label{sec:related}

\subsection{Efficient reasoning models}
LRMs allocate additional test-time computation to explicit reasoning traces often with specific delimiters such as \texttt{<think>}\,\ldots\,\texttt{</think>}~\cite{guo2025deepseek, jaech2024openai}. This \emph{think-before-answer} format enables iterative exploration and self-verification, but it is frequently applied \emph{uniformly} across inputs. Such uniform long-form reasoning leads to the \emph{overthinking} challenge with high latency and compute on intrinsically easy queries and sometimes producing redundant reasoning tokens \cite{feng2025efficient, sui2025stop}.

A broad line of work has pursued efficiency via \emph{adaptive computation} and \emph{early stopping}. Classical dynamic-computation methods and early-exit approaches terminate generation once confidence criteria are met, reducing unnecessary computation without changing the core model objective \cite{schuster2022confident, bae2025mixture}. Recent reasoning-focused extensions similarly aim to avoid excessive reasoning tokens by learning or heuristically determining when further context is unlikely to help, and by truncating or suppressing low-information reasoning segments \cite{sui2025stop}.

Another direction controls reasoning \emph{length} through prompt- or budget-based strategies. Methods that rely on fixed or predicted token budgets can reduce verbose CoT while maintaining accuracy, but often require external budget selection rules or calibrated estimators, which may not transfer cleanly across domains \cite{han2024token}. In parallel, internal-rationale approaches such as Quiet-STaR \cite{zelikman2024quiet} encourage models to form latent rationales during generation, while simplified recipes such as s1 \cite{muennighoff2025s1} show that carefully constructed 1k data and prompting can recover much of test-time scaling benefits with minimal training overhead.

As an orthogonal axis over these approaches, our goal is not to enforce a budget or modify the architecture, but to \emph{internalize the gating decision itself} whether to open a long explicit reasoning trace in the first place. \Ours~optimizes only the \emph{root token} that triggers the \texttt{<think>} branch, enabling adaptive reasoning with minimal training footprint and without backpropagating over whole reasoning trajectories.

\subsection{Difficulty-aware policy optimization}
The question of \textit{learning when to reason} has recently been reframed as \emph{difficulty-aware policy optimization}: the model should allocate explicit reasoning only when needed, while responding directly to easy or non-reasoning queries. This direction is especially relevant under the explicit \texttt{<think>} styled format, where the core idea is a discrete \emph{branching action} that enters a long reasoning mode versus produces a direct response \cite{Zhang2025AdaptThink, Bao2025Thinkless, wu2025arm, wang2025adareasoner}.

Representative methods learn such policies through RL or preference learning. AdaptThink \cite{Zhang2025AdaptThink} trains a policy to select \emph{thinking} versus \emph{non-thinking} modes conditioned on difficulty, while Thinkless \cite{Bao2025Thinkless} uses preference/RL signals to reduce redundant thinking tokens in general conversations. Beyond binary gating, ARM \cite{wu2025arm} expands the action space to multiple reasoning formats (e.g., direct answer, short CoT, long CoT, code), and AdaReasoner \cite{wang2025adareasoner} treats reasoning configurations as actions in a model-agnostic RL framework. Together, these works show that adaptive reasoning is learnable and can outperform static long-CoT inference.

However, existing policy optimization approaches face practical hurdles that motivate our design. 
First, many studies are benchmark-centric and largely evaluated on math-heavy suites, leaving open how well their learned routing \emph{calibrates} (e.g., whether the thinking rate remains aligned with difficulty on open-domain and scientific queries). 
Second, optimizing policies over whole long reasoning trajectories is often \emph{training-cost dominated}: GRPO-style post-training incurs expensive cost of VRAM, gradient computation, and  wall-clock costs \cite{guo2025deepseek}, also detailed in \S\ref{subsec:rpo} and \Autoref{tab:pflops_adaptive_reasoning_8b}. 
Third, effective training frequently depends on curated external prompts, data from larger teacher models, paired supervision, or proprietary query streams for routing signals, and these factors are rarely opened to the community.

\section{Method}
\label{sec:method}

\subsection{Motivation}
Adaptive reasoning is fundamentally a \emph{routing} problem. Before generating an answer, the model must decide whether to spend tokens on explicit deliberation.
In open-domain tasks, supervising this routing is difficult because task \textit{difficulty} is latent: correctness can be ambiguous and the same query may be easy or hard depending on a model's calibration and knowledge.
As a result, fixed heuristics (e.g., always long CoT or static budgets) are often miscalibrated.

\paragraph{Why math as a proxy}
We ground our training signal in mathematical reasoning because it admits \emph{deterministic} verification and a \emph{model-relative} notion of when reasoning pays off.
For each input query $x$, we elicit two mode-conditioned behaviors: (i) a direct-answer mode (\textsc{Short}) and (ii) an explicit-reasoning mode (\textsc{Long}). 
Let $c_S(x),c_L(x)\in\{0,1\}$ denote exact correctness under each mode. 
Rather than relying on external difficulty labels, we can use the difference between mode-conditioned success rates (e.g., Pass@$k$) as an empirical difficulty proxy: instances where \textsc{Long} is more reliable than \textsc{Short} are treated as harder for the current model, and vice versa.
This model-specific signal provides supervision for learning routing policies without external difficulty labels.

\begin{figure}
    \centering
    \includegraphics[width=0.95\linewidth]{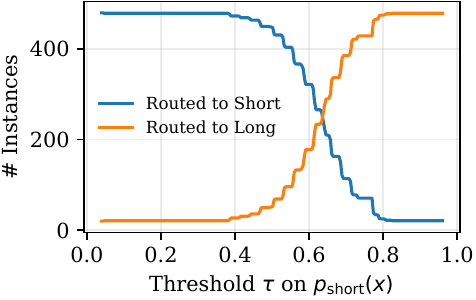}
      \caption{Thresholding on a \textsc{Short}-mode confidence signal yields a smooth routing trade-off about probability threshold vs number of routed instances over MATH500 dataset~\cite{hendrycks2021measuring}.}
    \label{fig:motivation-threshold}
\end{figure}

\paragraph{Self-taught difficulty signals}
After a lightweight SFT step that teaches the model to follow our \textsc{Short}/\textsc{Long} templates (\S\ref{subsec:sft}), we observe that routing can be anticipated at the generation root.
We obtain a model-specific difficulty signal by contrasting mode-conditioned accuracy on the same query (e.g., Pass@$k$ under \textsc{Short} vs.\ \textsc{Long}).
To expose this signal for routing, we \emph{parameterize} the mode decision at the generation root: under our template, the first newline token(s) after \texttt{<think>} select \textsc{Long} vs.\ \textsc{Short} (\texttt{\textbackslash n} vs.\ \texttt{\textbackslash n\textbackslash n}).
Let $z_S$ denote the \textsc{Short} root token (\texttt{\textbackslash n\textbackslash n}) and define a root-level \textsc{Short}-preference proxy as the model's conditional probability at this decision point:
\[
p_{\mathrm{short}}(x)\;=\;p_\theta\!\left(z_S \mid x,\texttt{<think>}\right).
\]
We route by thresholding:
\[
\pi_\tau(x)=
\begin{cases}
\textsc{Short} & \text{if } p_{\mathrm{short}}(x)\ge \tau,\\
\textsc{Long}  & \text{otherwise}.
\end{cases}
\]
Varying $\tau$ yields a simple calibration knob (\Autoref{fig:motivation-threshold}): larger $\tau$ makes the router more conservative, assigning fewer queries to \textsc{Short} and increasing the fraction routed to \textsc{Long}.
This synthetically produces an interpretable accuracy--routing trade-off, where intermediate thresholds preserve overall accuracy while reducing \textsc{Long} usage (\Autoref{tab:motivation-threshold}).

\begin{table}[t]
\centering
\small
\resizebox{\linewidth}{!}{%
\begin{tabular}{l|lll}
\toprule
Method ($\tau$) & \textsc{Short} & \textsc{Long} & Overall \\
\midrule
Original & 0.00 (0) & 83.6 (500) & 83.6 (500) \\
Adaptive (0.75) & 93.6 (62) & 82.4 (438) & 83.8 (500) \\
Adaptive (0.70) & 89.7 (155) & 78.8 (345) & 82.2 (500) \\
\bottomrule
\end{tabular}%
}
\caption{Accuracy (\%) and routed instance counts (in parentheses) under threshold routing over MATH500 dataset~\cite{hendrycks2021measuring} with R1-1.5B model.}
\label{tab:motivation-threshold}
\end{table}

\begin{figure}[t]
  \centering
  \begin{subfigure}[t]{0.48\linewidth}
    \centering
    \includegraphics[width=\linewidth]{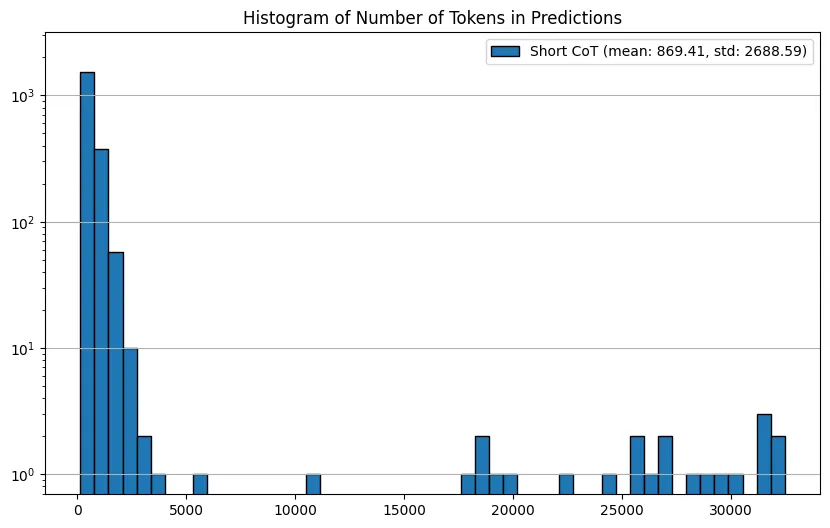}
    \caption{\textsc{Short}}
    \label{fig:tokenlen-short}
  \end{subfigure}\hfill
  \begin{subfigure}[t]{0.48\linewidth}
    \centering
    \includegraphics[width=\linewidth]{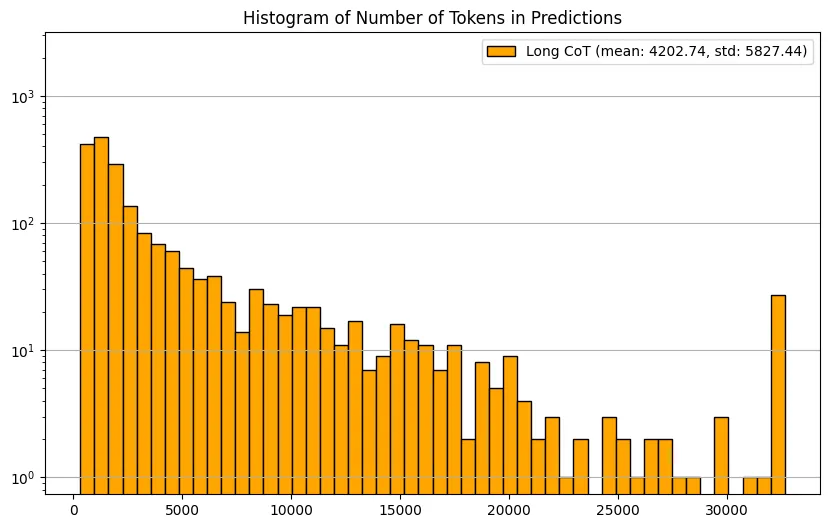}
    \caption{\textsc{Long}}
    \label{fig:tokenlen-long}
  \end{subfigure}\hfill
  \caption{Token-length histograms (log-scaled frequency) for \textsc{Short} and \textsc{Long} mode.}
  \label{fig:tokenlen}
\end{figure}

Beyond accuracy, routing is motivated by \emph{compute}. We measure generation length $L(x)$ (output tokens) and find that \textsc{Long} exhibits a heavy-tailed length distribution (\Autoref{fig:tokenlen}), i.e., a small fraction of queries trigger disproportionately long traces. This makes the routing decision directly translate into cost control, and motivates learning a lightweight gating policy at the root that reduces unnecessary \textsc{Long} usage without the loss of the model capability.

\subsection{SFT for mode selection}\label{subsec:sft}

\paragraph{Mode definition}
We define two generation modes using only the reasoning-tag template at the generation root.
The token \texttt{<think>} is treated as a special token, and \texttt{\textbackslash n} vs.\ \texttt{\textbackslash n\textbackslash n} are tokenized into different root tokens.
Concretely, we use:
\begin{itemize}
  \item \textsc{Long}: prefix \texttt{<think>\textbackslash n} so that the model continues with explicit reasoning content.
  \item \textsc{Short}: prefix \texttt{<think>\textbackslash n\textbackslash n</think>\textbackslash n\textbackslash n} to suppress explicit reasoning.
\end{itemize}

\begin{algorithm}[t]
\caption{SFT for mode selection}
\label{alg:sft}
\small
\begin{algorithmic}[1]
\Require base model $\pi_0$, prompts $\mathcal{D}$, samples per prompt $M$, tolerance $\epsilon$, budget $K$
\State $a_0 \gets \mathrm{Eval}(\pi_0;\ \textsc{DefaultChat})$ \Comment{default template}
\State $\pi \gets \pi_0$
\While{$\big|\mathrm{Eval}(\pi;\ \textsc{DefaultChat}) - a_0\big| \le \epsilon$}
  \State $\mathcal{S}_{S} \gets \mathrm{CollectCorrect}(\pi,\mathcal{D},M;\ \textsc{ShortPrefix})$
  \State $\mathcal{S}_{L} \gets \mathrm{CollectCorrect}(\pi,\mathcal{D},M;\ \textsc{LongPrefix})$
  \State $\mathcal{T} \gets \mathrm{Sample}(\mathcal{S}_{S},K)\ \cup\ \mathrm{Sample}(\mathcal{S}_{L},K)$
  \State $\pi \gets \mathrm{SFT}(\pi,\mathcal{T})$
\EndWhile
\end{algorithmic}
\end{algorithm}

\paragraph{Self-generated trajectories from OpenR1-Math queries}
We construct SFT supervision from OpenR1-Math queries by sampling multiple generations from the given LRM under each mode template (\textsc{Short}/\textsc{Long}).
We retain only generations that (i) satisfy the mode-specific template and (ii) match the exact answer, yielding two pools of correct demonstrations, $\mathcal{S}_{S}$ and $\mathcal{S}_{L}$.
We then randomly sample $K$ instances from each pool for SFT.
We do not enforce that a problem must be correct in both modes. 

\paragraph{SFT}
We use a SFT loop (\Autoref{alg:sft}) to manifest mode-conditioned \emph{style} without performance degradation under the default chat template.
Let $\mathrm{Acc}_{\mathrm{orig}}$ $=\mathrm{Eval}(\pi_0;\textsc{DefaultChat})$ be the initial accuracy, and let $\mathrm{Acc}(\pi)=\mathrm{Eval}(\pi;\textsc{DefaultChat})$ denote the same metric for an SFT candidate $\pi$.
We iteratively refresh the correctness-filtered SFT dataset and continue training only while the performance gap remains within a tolerance $\epsilon$:
\[
\big|\mathrm{Acc}(\pi) - \mathrm{Acc}_{\mathrm{orig}}\big| \le \epsilon.
\]
The loop is intentionally general. Empirically, across LRMs we tested, a single iteration typically sufficed (i.e., the constraint was satisfied without repeated resampling).

\subsection{Root-token policy optimization (\Ours)}\label{subsec:rpo}
SFT establishes \textsc{Short}/\textsc{Long} behaviors, but reliable \emph{adaptive selection} requires directly optimizing the gating policy.
The post-SFT model already knows \emph{how} to produce either style; what remains is learning \emph{when} to invoke each one.
We therefore view adaptive reasoning as a \emph{branching} problem: once the gate is chosen, the continuation is simply a mode-conditioned rollout rather than a learning target.
Motivated by this hypothesis, we apply \Ours~only to the first gating token (the root), treating the remainder of each sampled trajectory purely as environment output.

\begin{algorithm}[t]
\caption{Root-token policy optimization}
\label{alg:root-grpo}
\small
\begin{algorithmic}[1]
\Require policy $\pi_{\theta}$ (init from SFT), reference $\pi_{\text{ref}}$ (frozen), prompts $\mathcal{D}_{\text{rl}}$,
rollouts per prompt $n$ (even), context budget $S$, temperature $\tau$, KL weight $\beta$,
mode weights $(\alpha_S,\alpha_L)$
\State \textbf{Define gating tokens:} $z \in \{\texttt{\textbackslash n},\ \texttt{\textbackslash n\textbackslash n}\}$
\State \textbf{Define templates:} $\textsc{Short}(z{=}\texttt{\textbackslash n\textbackslash n})$, $\textsc{Long}(z{=}\texttt{\textbackslash n})$
\For{each training step}
  \State Sample a minibatch $\mathcal{B} \subset \mathcal{D}_{\text{rl}}$
  \For{each prompt $x \in \mathcal{B}$}
    \State \textbf{Balanced rollouts:} generate $n/2$ trajectories with \textsc{Short} template and $n/2$ with \textsc{Long}
    \State $\{(z_i^{(x)}, y_i^{(x)})\}_{i=1}^{n} \gets \mathrm{Rollout}(\pi_{\theta}, x;\ S,\tau)$ \Comment{$z_i^{(x)}$ is the root newline token(s)}
    \State Compute correctness indicators $c_i^{(x)} \gets \mathbf{1}\{\mathrm{Ans}(y_i^{(x)})\ \text{is correct}\}$
    \State Let $C_S(x) \gets \sum_{i:\,z_i^{(x)}=\texttt{\textbackslash n\textbackslash n}} c_i^{(x)},\quad C_L(x) \gets \sum_{i:\,z_i^{(x)}=\texttt{\textbackslash n}} c_i^{(x)}$
    \For{$i=1$ to $n$}
      \State $r_i^{(x)} \gets \alpha_S \cdot C_S(x)$ if $z_i^{(x)}=\texttt{\textbackslash n\textbackslash n}$ else $r_i^{(x)} \gets \alpha_L \cdot C_L(x)$
    \EndFor
    \State Store root-token log-probs $\ell_i^{(x)} \gets \log \pi_{\theta}(z_i^{(x)}\!\mid\! x)$ and $\ell_i^{(x),\text{ref}} \gets \log \pi_{\text{ref}}(z_i^{(x)}\!\mid\! x)$
  \EndFor
  \State \textbf{Per-mode group-relative advantages} :
  \State $\mathcal{I}_m \gets \{(x,i) : x \in \mathcal{B},\ m_i^{(x)} = m\}$ 
  \State $\mu_m \gets \frac{1}{|\mathcal{I}_m|}\!\!\sum_{(x,i) \in \mathcal{I}_m}\!\! r_i^{(x)},\quad
         \sigma_m^2 \gets \frac{1}{|\mathcal{I}_m|}\!\!\sum_{(x,i) \in \mathcal{I}_m}\!\!(r_i^{(x)}-\mu_m)^2$
  \State $A_i^{(x)} \gets (r_i^{(x)} - \mu_{m_i^{(x)}}) / (\sigma_{m_i^{(x)}} + \varepsilon)\quad \forall (x,i)$
  \State \textbf{Root-token-only update} (mask all tokens except $z$):
  \State $\theta \gets \mathrm{Update}\Big(\theta,\ \nabla_{\theta}\big[
    -\frac{1}{|\mathcal{B}|}\sum_{x\in\mathcal{B}}\frac{1}{n}\sum_{i=1}^{n} A_i^{(x)}\,\ell_i^{(x)}
    + \beta\cdot \mathrm{KL}(\pi_{\theta}(z\!\mid\!x)\ \Vert\ \pi_{\text{ref}}(z\!\mid\!x))
  \big]\Big)$
\EndFor
\end{algorithmic}
\end{algorithm}

\paragraph{Root-token action space}
We fix a \texttt{<think>} prefix and parameterize gating by the first newline tokenization:
\vspace{-10pt}
\[
z \in \{\texttt{\textbackslash n},\; \texttt{\textbackslash n\textbackslash n}\}.
\]
Because \texttt{<think>} is a special token and \texttt{\textbackslash n} vs.\ \texttt{\textbackslash n\textbackslash n} are distinct tokens, $z$ is a well-defined categorical action.
Selecting $z=\texttt{\textbackslash n}$ enters \textsc{Long}: the model continues generating inside the \texttt{<think>} block (explicit reasoning).
Choosing $z=\texttt{\textbackslash n\textbackslash n}$ enters \textsc{Short}: the template immediately closes the thinking block, i.e.,
\texttt{<think>\textbackslash n\textbackslash n</think>\textbackslash n\textbackslash n}.
\Ours~optimizes only $p_\theta(z\mid x)$; subsequent tokens are masked from the gradient.

\paragraph{Balanced rollouts}
For each query $x$, we collect $n$ rollouts with a fixed split: $n/2$ \textsc{Short} and $n/2$ \textsc{Long}.
Let $C_S(x)$ and $C_L(x)$ be the numbers of correct rollouts within the \textsc{Short} and \textsc{Long} groups, respectively.
This balanced sampling prevents trivial mode collapse and yields comparable advantage statistics across modes.

\begin{table*}[t]
\centering
\small
\resizebox{\linewidth}{!}{%
\begin{tabular}{lllllrr}
\toprule
Model & Method & Updated tokens $T$ & KV / rollout (GiB) & Act(bp) / rollout (GiB) & Train+Opt (GiB) & Peak @ $B{=}8$ (GiB) \\
\midrule
1.5B
& Original GRPO & 12K & 2.38 & 13.18 & 22.35 & 146.86 (1.00$\times$)\\
& \Ours & 1 & 2.38 & $1.10 \times 10^{-3}$ & 22.35 & 41.40 (0.28$\times$)\\
& \Ours~with LoRA update & 1 & 2.38 & $1.10 \times 10^{-3}$ & 0.07 & 19.12 (0.13$\times$)\\
\midrule
7B
& Original GRPO & 12K & 6.35 & 35.16 & 104.31 & 436.34 (1.00$\times$) \\
& \Ours & 1 & 6.35 & $2.93 \times 10^{-3}$ & 104.31 & 155.11 (0.36$\times$)\\
& \Ours~with LoRA update & 1 & 6.35 & $2.93 \times 10^{-3}$ & 0.21 & 51.01 (0.17$\times$)\\
\bottomrule
\end{tabular}%
}
\caption{
Peak VRAM (GiB) for long-context GRPO with uniform context $S{=}13$K (1K prompt${+}$12K generation) and $B{=}8$ concurrent rollouts. \Ours~gains memory efficiency by updating the root token ($T{=}1$).
LoRA further reduces trainable-state memory with fewer trainable parameters. 
}
\label{tab:vram-root-vs-vanilla}
\end{table*}

\paragraph{Prefix-level reward}
For a fixed prompt $x$, we refer to the $n/2$ rollouts that share a forced mode as the \emph{per-prompt mode group} (one for \textsc{Short}, one for \textsc{Long}). Each rollout is rewarded by the success count of its per-prompt mode group:
\vspace{-10pt}
\[
r_i^{(x)}=
\begin{cases}
\alpha_S \cdot C_S(x) & \text{if rollout } i \text{ is in \textsc{Short}},\\
\alpha_L \cdot C_L(x) & \text{if rollout } i \text{ is in \textsc{Long}}.
\end{cases}
\]
We set $\alpha_L{=}1$ and use a mild \textsc{Short} preference $\alpha_S{=}1+\frac{1}{n}$ to reflect the objective of saving compute without sacrificing the correctness.
Because $r_i^{(x)}$ depends on the per-prompt mode-group outcome, the update favors routing toward the mode that yields higher group-level correctness for that prompt.

\paragraph{Per-mode group-relative advantages}
To optimize \emph{switching} under the enforced split, we standardize advantages within each mode by pooling rewards. Per-prompt-mode would be degenerate: by construction, $r_i^{(x)}=\alpha_{m_i}\!\cdot\! C_{m_i}(x)$ is constant across the $n/2$ rollouts that share a mode within prompt $x$, so the within-mode-group variance is identically zero. Cross-prompt variation in $C_S(x)$ and $C_L(x)$ is what supplies the non-degenerate variance needed for advantage estimation. Concretely, for rollout $i$ of prompt $x$ in mode $m_i^{(x)} \in \{S, L\}$, let $\mathcal{I}_m = \{(x,i) : x \in \mathcal{B},\ m_i^{(x)} = m\}$ denote the \emph{group-level mode pool}. Then,

\[
\resizebox{\linewidth}{!}{$
\begin{aligned}
A_i^{(x)} &= \frac{r_i^{(x)} - \mu_{m_i^{(x)}}}{\sigma_{m_i^{(x)}} + \varepsilon}, \\
\mu_m &= \frac{1}{|\mathcal{I}_m|}\!\!\sum_{(x,j) \in \mathcal{I}_m}\!\! r_j^{(x)},\quad
\sigma_m^2 = \frac{1}{|\mathcal{I}_m|}\!\!\sum_{(x,j) \in \mathcal{I}_m}\!\!\bigl(r_j^{(x)}-\mu_m\bigr)^2.
\end{aligned}
$}
\]

Pooling is \emph{per-mode} (separate $\mu_S, \sigma_S$ and $\mu_L, \sigma_L$) so that the standardized advantages remain on a comparable scale even when the two modes have systematically different success rates; pooling is \emph{group-level} so that within-mode variance is non-zero. The signal each $A_i^{(x)}$ encodes is therefore ``how does prompt $x$'s mode-$m$ success compare to the group-wide mode-$m$ baseline,'' which the policy converts into an updated routing distribution over $z$. This per-mode group relative scheme presupposes prompt-difficulty diversity within $\mathcal{B}$.

\begin{table*}[t!]
\centering
\small
\setlength{\tabcolsep}{4pt}
\renewcommand{\arraystretch}{1.1}
\resizebox{\linewidth}{!}{%
\begin{tabular}{llrrrrrrrrr|rrrr}
\toprule
Model & Stage
& \multicolumn{3}{c}{MATH500}
& \multicolumn{3}{c}{AIME24}
& \multicolumn{3}{c}{AIME25}
& \multicolumn{4}{|c}{Avg} \\
\cmidrule(lr){3-5}\cmidrule(lr){6-8}\cmidrule(lr){9-11}\cmidrule(lr){12-15}
 &  & P@1 & \#Tok & Think\% & P@1 & \#Tok & Think\% & P@1 & \#Tok & Think\%
 & P@1 & \#Tok & Think\% & RAR$\downarrow$ \\
\midrule
Qwen3-1.7B & Baseline (\textsc{Long})
& 91.1 & 4,539 & 100.0 & 45.6 & 15,569 & 100.0 & 38.1 & 15,830 & 100.0
& 58.3 & 11,979 & 100.0 & 1.72 \\
& Baseline (\textsc{Short})
& 73.0 & 978 & 0.0 & 12.3 & 3,898 & 0.0 & 11.5 & 2,180 & 0.0
& 32.3 & 2,352 & 0.0 & 0.00 \\
 & + SFT
& 85.5 & 3,767 & 69.8 & 34.4 & 12,558 & 72.9 & 30.4 & 12,648 & 75.6
& 50.1 & 9,658 & 72.8 & 1.45 \\
 & \cellcolor{gray!20}+ \Ours
& \cellcolor{gray!20} 79.5 & \cellcolor{gray!20} 2,754 & \cellcolor{gray!20} 31.8 & \cellcolor{gray!20} 43.8 & \cellcolor{gray!20} 13,315 & \cellcolor{gray!20} 93.8 & \cellcolor{gray!20} 31.7 & \cellcolor{gray!20} 14,231 & \cellcolor{gray!20} 96.3
& \cellcolor{gray!20} 51.7 & \cellcolor{gray!20} 10,100 & \cellcolor{gray!20} 74.0 & \cellcolor{gray!20} 1.43 \\
\midrule
Qwen3-8B & Baseline (\textsc{Long})
& 96.6 & 4,777 & 100.0 & 74.4 & 14,184 & 100.0 & 64.8 & 16,950 & 100.0
& 78.6 & 11,970 & 100.0 & 1.27 \\
& Baseline (\textsc{Short})
& 84.1 & 1,118 & 0.0 & 25.4 & 5,646 & 0.0 & 20.8 & 3,159 & 0.0
& 43.4 & 3,308 & 0.0 & 0.00 \\
 & + SFT
& 90.6 & 3,390 & 53.5 & 54.0 & 12,350 & 59.2 & 46.7 & 12,661 & 59.2
& 63.8 & 9,467 & 57.3 & 0.90 \\
 & \cellcolor{gray!20}+ \Ours
& \cellcolor{gray!20} 90.5 & \cellcolor{gray!20} 2,794 & \cellcolor{gray!20} 25.0 & \cellcolor{gray!20} 52.9 & \cellcolor{gray!20} 12,784 & \cellcolor{gray!20} 56.4 & \cellcolor{gray!20} 42.5 & \cellcolor{gray!20} 12,029 & \cellcolor{gray!20} 53.1
& \cellcolor{gray!20} 62.0 & \cellcolor{gray!20} 9,202 & \cellcolor{gray!20} 44.8 & \cellcolor{gray!20} 0.72 \\
\midrule
Exaone4-1.2B & Baseline (\textsc{Long})
& 91.0 & 5,416 & 100.0 & 54.0 & 18,529 & 100.0 & 43.8 & 19,124 & 100.0
& 62.9 & 14,356 & 100.0 & 1.59 \\
& Baseline (\textsc{Short})
& 75.6 & 854 & 0.0 & 21.9 & 3,075 & 0.0 & 15.2 & 2,718 & 0.0
& 37.6 & 2,216 & 0.0 & 0.00 \\
 & + SFT
& 84.6 & 3,015 & 40.9 & 37.3 & 9,524 & 39.8 & 31.3 & 10,562 & 49.4
& 51.1 & 7,700 & 43.4 & 0.85 \\
 & \cellcolor{gray!20}+ \Ours
& \cellcolor{gray!20} 83.5 & \cellcolor{gray!20} 2,455 & \cellcolor{gray!20} 25.5 & \cellcolor{gray!20} 33.8 & \cellcolor{gray!20} 6,592 & \cellcolor{gray!20} 21.9 & \cellcolor{gray!20} 27.3 & \cellcolor{gray!20} 6,665 & \cellcolor{gray!20} 23.3
& \cellcolor{gray!20} 48.2 & \cellcolor{gray!20} 5,237 & \cellcolor{gray!20} 23.6 & \cellcolor{gray!20} 0.49 \\
\bottomrule
\end{tabular}%
}
\caption{Baseline $\rightarrow$ SFT mode selection $\rightarrow$ \Ours~on math benchmarks.
Pass@1 (P@1) is the average accuracy (\%) over 16 runs. \#Tok is average generated tokens. Think\% is the fraction routed to \textsc{Long}.
\textbf{Avg} columns are simple averages over (MATH500, AIME24, AIME25).
\textbf{RAR} (\(\downarrow\)) is the reasoning-per-accuracy ratio, defined as
\(\mathrm{RAR}_{\text{avg}} = \frac{t_{\text{avg}}}{a_{\text{avg}}+\varepsilon}\), where \(a_{\text{avg}}\) and \(t_{\text{avg}}\) are Avg Pass@1 and Avg Think\% as fractions. Lower is better.}
\label{tab:main_math_progress}
\vspace{-10pt}
\end{table*}

\paragraph{\Ours~objective with KL regularization}
We regularize the log-probability of the root action $z$ with KL-to-reference loss:
\begin{align}
\max_{\theta}\;\; & \mathbb{E}_{x}\!\left[\frac{1}{n}\sum_{i=1}^{n} A_i\,\log p_{\theta}(z_i\!\mid x)\right] \nonumber\\
&-\beta\;\mathrm{KL}\!\left(p_{\theta}(z\!\mid x)\,\Vert\,p_{\theta_{\mathrm{ref}}}(z\!\mid x)\right).
\end{align}
where $\theta_{\mathrm{ref}}$ is the post-SFT checkpoint.
In implementation, we generate full trajectories to evaluate correctness and compute rewards, but retain only the root log-probabilities. Other non-root tokens are discarded from the training graph, and the update is performed solely on $p_{\theta}(z{\mid}x)$.

\paragraph{Balanced difficulty sampling}
We perform \Ours~on a compact set of $\sim$1K queries comprised of a part of s1k \cite{muennighoff2025s1} and compression dataset \cite{arora2025training}.
With the post-SFT model, We estimate mode-conditioned success rates and exclude items where \textsc{Long} is rarely correct ($\mathrm{acc}_{L}{<}\tau_{\min}{=}0.25$).
We then stratify the remaining queries into \textit{easy} and \textit{hard} regimes and sample them in a 1:1 ratio to stabilize updates and avoid mode collapse.
Concretely, \textit{easy} includes queries where \textsc{Short} is already competitive (e.g., $\mathrm{acc}_{S}{\ge} \tau_{\text{easy}}{=}0.75$), while \textit{hard} includes queries where \textsc{Long} is consistently preferable (e.g., $\mathrm{acc}_{S}{<}\mathrm{acc}_{L}{\le} \tau_{\text{hard}}$ with $\tau_{\text{hard}}{=}0.5$).

\paragraph{Why root-only updates are memory-efficient} 
Both vanilla GRPO and \Ours\ must work with KV-cache over the full rollout context length $S$ during sampling. The training-time gap arises from backpropagation. Vanilla GRPO updates all generated tokens, whereas \Ours\ masks the trajectory and updates only the root gating token ($T{=}1$). As a result, the dominant activation term scales with the number of updated tokens,
\[
\mathcal{M}_{\text{bp}} \propto B \cdot T \cdot L \cdot d \cdot b \cdot \kappa,
\]
where $B$ is the number of rollouts processed concurrently, $L$ and $d$ are model depth and width, $b$ is the activation precision (e.g., bf16), and $\kappa$ aggregates saved intermediates (e.g., attention/MLP). Thus, setting $T{=}1$ removes the linear dependence on long CoT length.
\Autoref{tab:vram-root-vs-vanilla} quantifies this effect under long-context rollout scenarios, where activation memory dominates the feasibility boundary.

\paragraph{Training configuration}
Unless noted otherwise, we use three H100 GPUs in a node.
Each iteration samples 8 queries and collects $n{=}16$ rollouts per query (8 per mode) with temperature 0.6 and top-$p$ 0.95. We allow maximum context length 16,384 for rollouts.

\section{Experiment}\label{sec:exp}

\subsection{Setup}
Our pipeline follows \textbf{Baseline $\rightarrow$ SFT $\rightarrow$ \Ours}, and we report (i) Pass@1 accuracy, (ii) average generated tokens, and (iii) the fraction routed to \textsc{Long} (Think\%) in \Autoref{tab:main_math_progress}.
We report results on diverse models with different families and sizes: \textbf{Qwen3-1.7B}, \textbf{Qwen3-8B}, \textbf{Qwen3-32B} \cite{yang2025qwen3}, \textbf{Exaone4-1.2B} \cite{research2025exaone} and \textbf{R1-1.5B} \cite{guo2025deepseek}.
All models use a \texttt{<think>} style template and share the same two-mode gating mechanism implemented via the first newline token(s) after \texttt{<think>}.

\begin{figure}[t]
  \centering
  \includegraphics[width=0.88\linewidth]{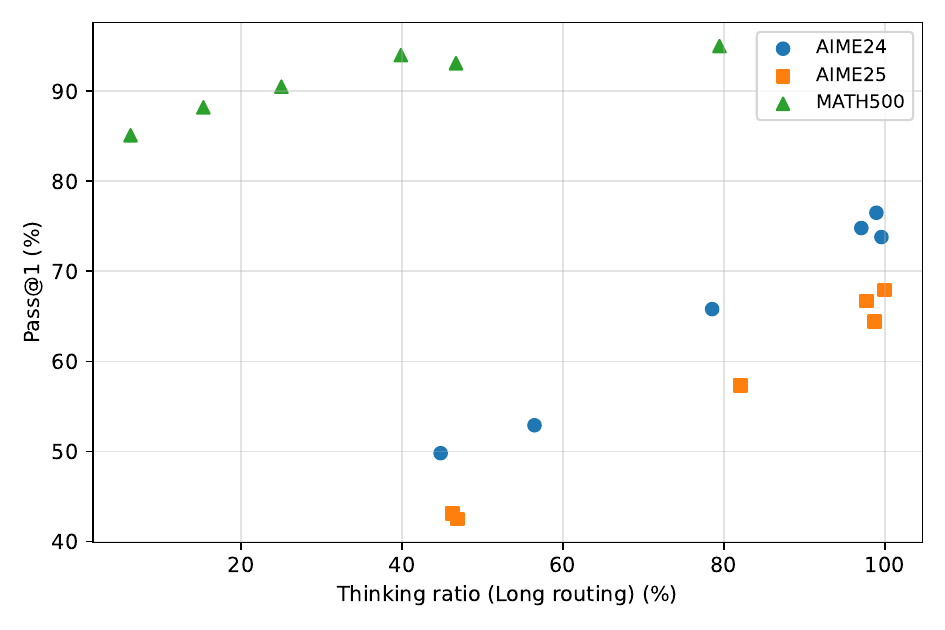}
  \caption{Think\% versus Pass@1 on Qwen3-8B with \Ours.
  Each point indicates an intermediate checkpoint.
  }
  \label{fig:calib-scatter}
  \vspace{-10pt}
\end{figure}

\subsection{Datasets and evaluation}
\paragraph{Math benchmarks}
We evaluate on three standard math reasoning sets: MATH500 \cite{hendrycks2021measuring}, {AIME24} \cite{maa_aime_2024}, and {AIME25} \cite{maa_aime_2025}, scientific reasoning dataset as GPQA \cite{rein2024gpqa}, and coding task with LiveCodeBench\footnote{LiveCodeBench is sliced from March 25 2025 to April 15 2025}.
Following prior work, correctness is determined by exact match on the final answer parsed by \texttt{boxed\{\}}. At inference time, we match the original technical reports including temperature and sampling settings.

\paragraph{SFT data for mode selection}
We use mathematical queries from OpenR1-Math data \cite{openr1_openr1_math_220k_2025} and self-generate candidate trajectories with the given model under each mode template.
We filter candidates by correctness and basic template validity, producing two pools of correct demonstrations for \textsc{Short} and \textsc{Long}.
We then sample 20K instances per mode (40K total) for SFT, without requiring overlap between the two pools.

\paragraph{Implementation details}
We implement SFT and \Ours~in \textbf{OpenRLHF} \cite{hu2024openrlhf}, with a customized lightweight \emph{asynchronous} rollout engine that overlaps generation and post-processing.
Unless noted, we use max context \textbf{16K}, batch size \textbf{128}, and learning rate \textbf{$10^{-6}$}.
For \Ours, we sample \textbf{16} rollouts per prompt (8/8 \textsc{Short}/\textsc{Long}) and apply gradients only to the root gating token.
We run $<\textbf{60}$ update steps for all model sizes.

\begin{table}[t]
\centering
\small
\setlength{\tabcolsep}{5pt}
\renewcommand{\arraystretch}{1.12}
\resizebox{\linewidth}{!}{%
\begin{tabular}{lll|ll|ll}
\toprule
Method
& \multicolumn{2}{c}{MATH500}
& \multicolumn{2}{c}{AIME24}
& \multicolumn{2}{c}{Avg}\\
\cmidrule(lr){2-3}\cmidrule(lr){4-5}\cmidrule(lr){6-7}
 & P@1 & Think\% & P@1 & Think\% & P@1 & RAR$\downarrow$ \\
\midrule
\multicolumn{7}{l}{\textit{Prior hybrid-reasoning baselines (reported in their tables)}} \\
Original (\textsc{Long})  & 83.2 & 100.0 & 29.4 & 100.0 & 56.3 & 1.78 \\
Original (\textsc{Short}) & 67.2 & 0.0 & 14.0 & 0.0 & 40.6 & 0.00 \\
Rejection Finetuning $^{\dagger}$ \cite{Zhang2025AdaptThink} & 72.4 & 66.6 & 25.2 & 79.0 & 48.8 & 1.49 \\
\cellcolor{gray!20}AdaptThink$^{\dagger}$ \cite{Zhang2025AdaptThink}  & \cellcolor{gray!20}82.0 & \cellcolor{gray!20}23.2 & \cellcolor{gray!20}31.0 & \cellcolor{gray!20}59.6 & \cellcolor{gray!20}56.5 & \cellcolor{gray!20}0.74 \\
Thinkless$^{\dagger}$ \cite{Bao2025Thinkless} & 81.9 & 51.6 & 27.3 & 100.0 & 54.6 & 1.39 \\
\midrule
\multicolumn{7}{l}{\textit{Ours (same backbone family; evaluated with our protocol)}} \\
+ SFT & 75.1 & 37.8 & 20.8 & 68.5 & 48.0 & 1.11 \\
\cellcolor{gray!20}+ \Ours & \cellcolor{gray!20}77.6 & \cellcolor{gray!20}20.5 & \cellcolor{gray!20}25.0 & \cellcolor{gray!20}77.7 & \cellcolor{gray!20}51.3 & \cellcolor{gray!20}0.96 \\
\bottomrule
\end{tabular}%
}
\caption{Comparison with accuracy and Think\% over R1-1.5B model. $\dagger$ presents the performance from the original paper. 
}\vspace{-10pt}
\label{tab:cherry_math500_aime24_acc_think}
\end{table}

\begin{table*}[t]
\centering
\small
\setlength{\tabcolsep}{5pt}
\renewcommand{\arraystretch}{1.12}
\resizebox{\linewidth}{!}{%
\begin{tabular}{lrrrrrr}
\toprule
Method 
& SFT spec
& GRPO spec
& Forward PFLOPs
& Backward PFLOPs
& Total PFLOPs
& Rel.\ Total (vs.\ AdaptThink) \\
\midrule
AdaptThink \cite{Zhang2025AdaptThink}
& 40K @ 16K
& 40K @ 16K, $K{=}16$, $T_{\text{upd}}{\approx}16$K
& 174{,}080
& 348{,}160
& 522{,}240
& 1.00$\times$ \\
\midrule
ThinkLess \cite{Bao2025Thinkless}
& 334K @ 24K
& 1K @ 24K, $K{=}8$, $T_{\text{upd}}{\approx}24$K
& 131{,}328
& 262{,}656
& 393{,}984
& 0.75$\times$ \\\midrule
\Ours~(Ours)
& 8K @ 16K
& 1K @ 16K, $K{=}16$, $T_{\text{upd}}{=}1$
& 6{,}144
& 4{,}097
& 10{,}241
& 0.02$\times$ \\
\bottomrule
\end{tabular}%
}
\caption{PFLOPs accounting for an 8B dense Transformer (full-length sequences at the stated context). We assume $2P$ FLOPs/token for forward and $4P$ for backward. For AdaptThink, the number of dataset is set to 40K as reported.}
\label{tab:pflops_adaptive_reasoning_8b}
\vspace{-10pt}
\end{table*}

\begin{table}[t]
\centering
\small
\setlength{\tabcolsep}{3.5pt}
\renewcommand{\arraystretch}{1.1}
\resizebox{\linewidth}{!}{%
\begin{tabular}{llrrrrrr|rrrr}
\toprule
Model & Stage
& \multicolumn{3}{c}{LiveCodeBench}
& \multicolumn{3}{c}{GPQA-Diamond}
& \multicolumn{4}{c}{Avg} \\
\cmidrule(lr){3-5}\cmidrule(lr){6-8}\cmidrule(lr){9-12}
 &  & P@1 & \#Tok & Think\% & P@1 & \#Tok & Think\%
 & P@1 & \#Tok & Think\% & RAR$\downarrow$ \\
\midrule
Exaone4-1.2B
& Baseline   & 42.2 & 12,159 & 100.0 & 50.8 &  7,145 & 100.0 & 46.5 &  9,652 & 100.0 & 2.15 \\
& + SFT      & 34.6 &  9,281 &  90.1 & 46.2 &  5,037 &  70.1 & 40.4 &  7,159 &  80.1 & 1.98 \\
& \cellcolor{gray!20} + \Ours    & \cellcolor{gray!20} 36.7 &  \cellcolor{gray!20} 6,692 &  \cellcolor{gray!20} 61.7 & \cellcolor{gray!20} 41.7 &  \cellcolor{gray!20} 2,881 &  \cellcolor{gray!20} 35.0 & \cellcolor{gray!20} 39.2 & \cellcolor{gray!20} 4,787 & \cellcolor{gray!20} 48.4 & \cellcolor{gray!20} 1.26 \\
\midrule
Qwen3-8B
& Baseline   & 49.5 & 14,878 & 100.0 & 58.8 &  7,355 & 100.0 & 54.2 & 11,117 & 100.0 & 1.85 \\
& + SFT      & 30.2 & 10,410 &  63.0 & 31.1 &  5,227 &  50.1 & 30.7 &  7,819 &  56.6 & 1.85 \\
& \cellcolor{gray!20} + \Ours    & \cellcolor{gray!20} 35.4 & \cellcolor{gray!20} 11,532 &  \cellcolor{gray!20} 54.7 & \cellcolor{gray!20} 55.4 & \cellcolor{gray!20} 6,910 & \cellcolor{gray!20} 67.2 & \cellcolor{gray!20} 45.4 & \cellcolor{gray!20} 9,221 & \cellcolor{gray!20} 61.0 & \cellcolor{gray!20} 1.34 \\
\midrule
Qwen3-32B $\dagger$  &
Baseline         &  59.1 & 13,498 & 100.0 & 66.2 &  5,108 & 100.0 & 62.7 & 9,303 & 100.0 & 1.60 \\
 & \cellcolor{gray!20} + SFT & \cellcolor{gray!20} 58.3 & \cellcolor{gray!20} 11,457 &  \cellcolor{gray!20} 90.6 & \cellcolor{gray!20} 66.3 & \cellcolor{gray!20} 4,822 & \cellcolor{gray!20} 98.5 & \cellcolor{gray!20} 62.3 & \cellcolor{gray!20} 8,140 & \cellcolor{gray!20} 94.6 & \cellcolor{gray!20} 1.52 \\
 & + \Ours       &  54.4 & 10,623 &  87.2 & 65.0 &  4,768 &  97.6 & 59.7 & 7,696 & 92.4 &  1.55 \\
\bottomrule
\end{tabular}%
}
\caption{LiveCodeBench/GPQA-Diamond results. Models without $\dagger$ are trained with full-parameter SFT; $\dagger$ indicates a LoRA-tuned model. 
}
\vspace{-10pt}
\label{tab:non_math_tasks}
\end{table}

\subsection{Main results}
\autoref{tab:main_math_progress} reports math performance across stages (\textbf{Baseline $\rightarrow$ SFT $\rightarrow$ \Ours}).
To summarize \emph{reasoning efficiency} in one number, we use \textbf{RAR} (\(\downarrow\)), the reasoning-per-accuracy ratio:
$\mathrm{RAR}_{\text{avg}}=\frac{t_{\text{avg}}}{a_{\text{avg}}+\varepsilon}$,
where \(a_{\text{avg}}\) is Avg Pass@1 and \(t_{\text{avg}}\) is Avg Think\% (both as fractions). %

\paragraph{\Ours~improves the accuracy routing trade-off}
On the larger and mid-scale backbones, \Ours~reduces Think\% with modest accuracy changes, yielding the best (lowest) RAR.
For example, on Qwen3-8B, Avg Think\% decreases from 57.3\% to 44.8\% while Avg Pass@1 stays close (63.8\%$\rightarrow$62.0\%), improving RAR from 0.90 to 0.72; on Exaone4-1.2B, Think\% drops from 43.4\% to 23.6\% and RAR improves from 0.85 to 0.49. 
On Qwen3-1.7B, the post-SFT operating point is already conservative on the LONG side (Think\%${=}72.8$, RAR${=}1.45$); \Ours~yields a comparable operating point (Think\%${=}74.0$, RAR${=}1.43$) that recovers accuracy via routing refinement rather than via further Think\% compression. 

\paragraph{SFT already enables routing; \Ours~refines it.}
SFT alone yields a strong reduction in Think\% (and tokens), showing that mode-following is largely learned at the template level.
\Ours~then fine-tunes the \emph{root} decision to better match query difficulty. 
The direction of this refinement is configuration-dependent: on Qwen3-1.7B and R1-1.5B (\Autoref{tab:cherry_math500_aime24_acc_think}), \Ours~recovers accuracy on top of SFT by reallocating LONG toward the harder subset; on Qwen3-8B and Exaone4-1.2B, \Ours~instead trades a small accuracy reduction for further Think\% compression.

\paragraph{Comparison with other GRPO-based methods}
\Autoref{tab:cherry_math500_aime24_acc_think} compares \Ours~with recent hybrid reasoning baselines on the same R1 1.5B model. \Ours~lowers the fraction routed to \textsc{Long} to 20.5\% and reaches a two-task average accuracy of 51.3 with RAR${=}0.96$. AdaptThink achieves a stronger absolute operating point (Avg P@1${=}56.5$, RAR${=}0.74$) but at a far larger training budget (\Autoref{tab:pflops_adaptive_reasoning_8b}); the comparison should therefore be read as two distinct points on a budget-accuracy frontier rather than a head-to-head ranking.
Additionally, \Autoref{tab:pflops_adaptive_reasoning_8b} highlights the compute-efficiency gap. Trajectory level GRPO is compute heavy because gradients traverse the full chain of thought. With matched accounting on an 8B reference model as an example, \Ours~reduces total training compute to about 0.02$\times$ AdaptThink.

\section{Analysis}\label{sec:Analysis}

\paragraph{Controlling the degree of routing}
\Autoref{fig:calib-scatter} plots P@1 against Think\% for intermediate \Ours~checkpoints. Points with comparable accuracy but different thinking ratios indicate multiple feasible operating points.
Thus, calibration is practical: choose a checkpoint that satisfies a compute/routing budget while maintaining accuracy. The achievable frontier depends on the RL prompt set and easy/hard mix, so checkpoint selection is dataset-conditioned rather than universal.

\paragraph{LiveCodeBench and GPQA}
\Autoref{tab:non_math_tasks} extends evaluation beyond math to coding and scientific QA, where the optimal amount of thinking is more heterogeneous.
Across models, SFT already introduces a meaningful reduction in explicit reasoning, while \Ours~further reshapes the accuracy routing profile by updating only the root decision.
The effect is model-size dependent: on Qwen3-8B and Exaone4-1.2B, \Ours~compresses average decoding effort and lowers RAR; on Qwen3-32B, where SFT already produces a near-baseline operating point (Avg Think\%${=}94.6$, RAR${=}1.52$), \Ours~yields a comparable but not strictly better RAR (1.55), reflecting that the room for further routing-only gains shrinks once the underlying solver is sufficiently strong and the SFT operating point is already conservative. Practically, this suggests treating the post-\Ours~checkpoint as a controllable router (with $\tau$-sweeps) rather than as a fixed point that uniformly dominates SFT.

\begin{figure}[t]
    \centering
    \begin{subfigure}[t]{0.49\linewidth}
        \centering
        \includegraphics[width=\linewidth]{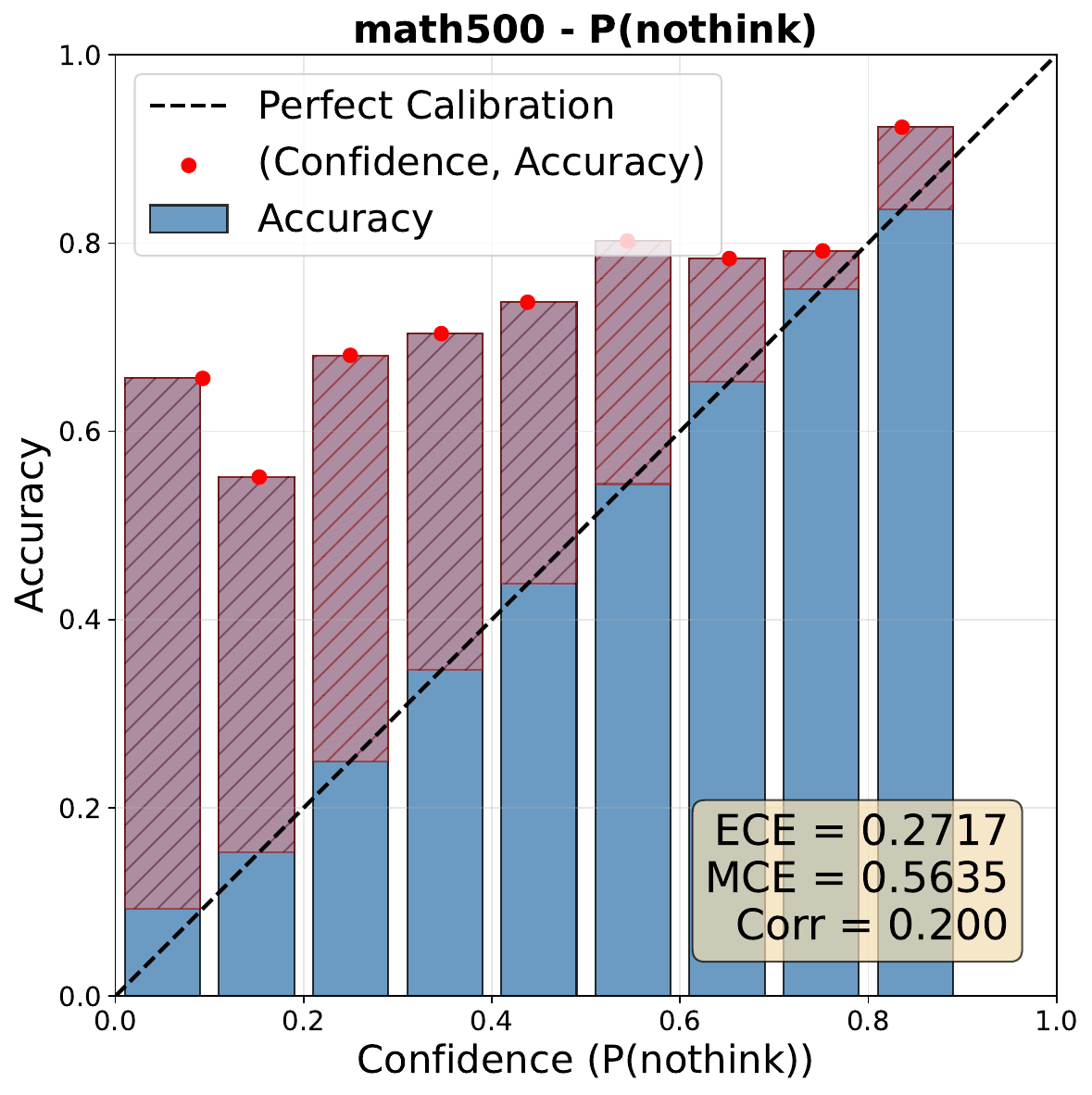}
        \caption{\textsc{Math500}.}
        \label{fig:calib-math500-nothink}
    \end{subfigure}
    \hfill
    \begin{subfigure}[t]{0.49\linewidth}
        \centering
        \includegraphics[width=\linewidth]{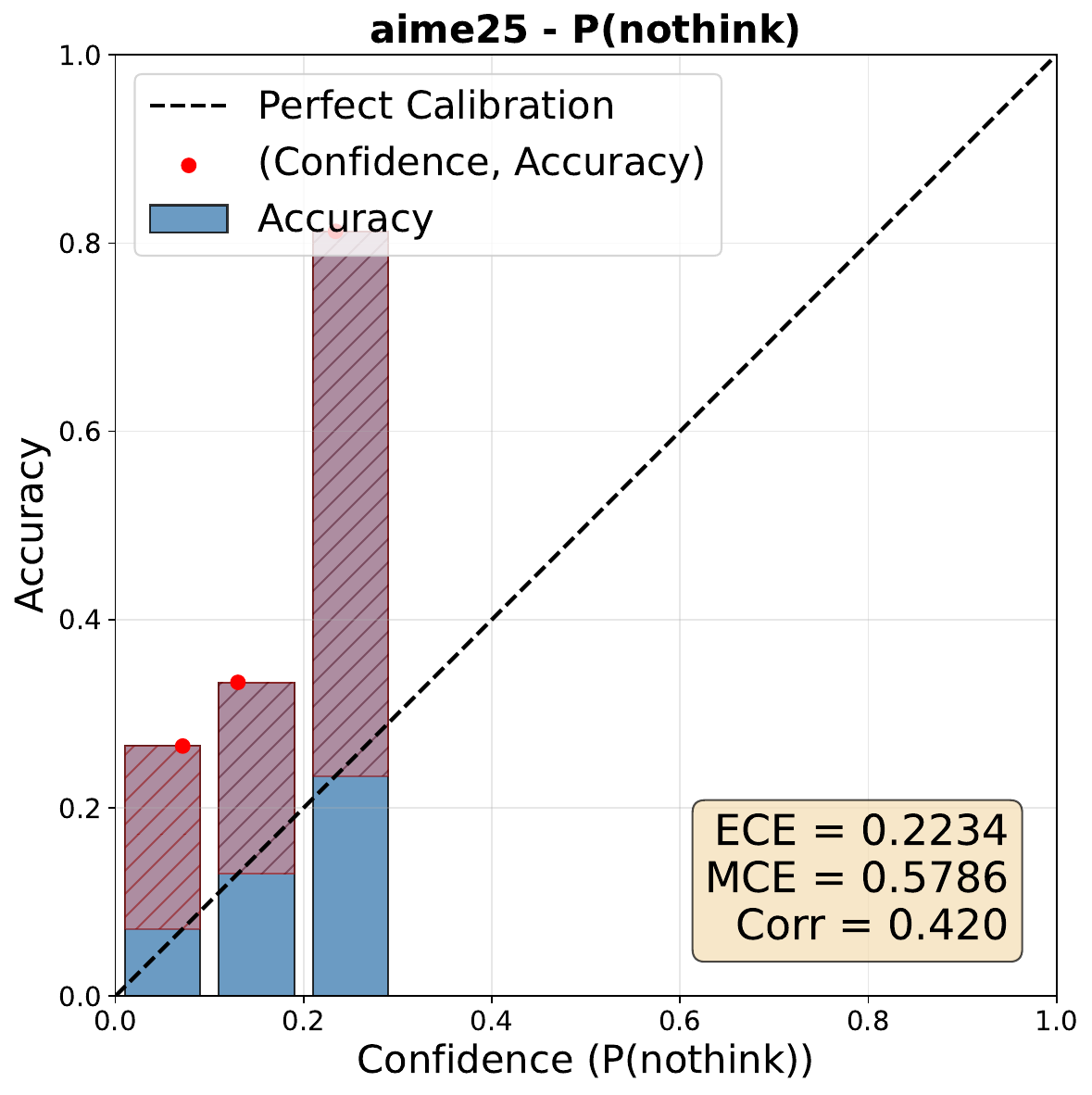}
        \caption{\textsc{AIME25}.}
        \label{fig:calib-aime25-nothink}
    \end{subfigure}
    \caption{Reliability diagrams \cite{guo2017calibration} for the root confidence $p_{\mathrm{short}}(x){=}p_\theta(z_S {\mid} x,\texttt{<think>})$. Each bin plots empirical accuracy against predicted confidence, with the diagonal indicating perfect calibration.}
    \label{fig:calib-nothink}
    \vspace{-10pt}
\end{figure}

\paragraph{Calibration of root confidence}
\Autoref{fig:calib-nothink} reports reliability diagrams with Qwen3-1.7B model for the root \textsc{Short} confidence $p_{\mathrm{short}}(x){=}p_\theta(z_S{\mid}x,\texttt{<think>})$. We bin $p_{\mathrm{short}}$ into 10 uniform intervals and compute (i) \textbf{Corr}, the Pearson correlation between per query confidence and empirical accuracy, and (ii) calibration errors \textbf{ECE} $=\sum_b \frac{n_b}{N}\lvert \mathrm{acc}_b-\mathrm{conf}_b\rvert$ and \textbf{MCE} $=\max_b \lvert \mathrm{acc}_b-\mathrm{conf}_b\rvert$, where $\mathrm{acc}*b$ and $\mathrm{conf}*b$ are the average accuracy and confidence within bin $b$ \cite{guo2017calibration}. Overall, $p_{\mathrm{short}}(\cdot)$ tends to correlate with correctness, but the signal is weak and dataset dependent. On AIME25, which is substantially harder, mass concentrates at low $p_{\mathrm{short}}(\cdot)$, so the router rarely assigns high \textsc{Short} confidence, whereas MATH500 shows a broader spread with more high confidence bins. 
Further discussions are in Appendix.

\vspace{-5pt}
\section{Conclusion}\label{sec:conclusion}
We introduce \Ours, a root token policy optimization framework for adaptive thinking that updates only the initial gating token while treating the remainder of the rollout as environment output. Precisely, \Ours~turns static \textsc{Long} style generation into a lightweight routing problem, reducing significant training overhead during GRPO.
Across multiple model families and scales, \Ours~improves the accuracy routing trade-off under most scenarios, achieving lower reasoning per accuracy ratios over expert scientific reasoning and coding as well as math reasoning. 
Some configurations trace a comparable but not strictly dominating frontier, supporting the use of \Ours~as a controllable routing layer rather than a uniform replacement for SFT.

\vspace{-5pt}
\section*{Limitations}
While our work supports diverse LRM families and scales, several limitations suggest avenues for future exploration. First, broader evaluation is needed under advanced benchmark datasets requiring agentic behaviors with tools.
Second, while root confidence is informative under our branching problem setting, it can be not only sensitive to bucket composition (i.e., easy and hard queries for \Ours), but weakly calibrated towards its query difficulty.
Third, we focus on a binary routing decision and do not optimize intra mode behaviors such as how long to think or when to stop, which could further improve compute allocation.
Fourth, \Ours's gain over the SFT operating point is configuration-dependent: on Qwen3-1.7B, the post-SFT routing already saturates the achievable RAR, so \Ours~yields a comparable rather than strictly better operating point; on Qwen3-32B, where we use LoRA tuning and SFT already produces a near-baseline conservative router, \Ours~likewise traces a comparable but not strictly dominating frontier. We therefore position \Ours~as a controllable routing layer to be selected via checkpoint or threshold, rather than a uniform replacement for SFT.
Finally, due to limited compute, we do not provide a more exhaustive ablation of the RL training dynamics regarding policy optimization to uncover how \Ours~enables such gating behaviors in model aspect or a fully controlled comparison against alternative GRPO variants based on their publicly released checkpoints and exact training recipes.
While mitigating such behaviors would be valuable, it is beyond the scope of this study. We leave other discussion points including the design of root token choice, and generalizability in Appendix.

\section*{Acknowledgments}

We thank Dahyun Lee for providing extensive feedback on our paper.

\bibliography{ACL2026/ref}

\newpage
\appendix

\section{Discussions}

\subsection{Root-level gating is tokenization agnostic}

We demonstrate that the design of root token in \Ours~framework is tokenization agnostic over a broader range of model families.
Other than the settings in the body, we observe that models without specific routing tokens  can achieve robust mode routing by prepending arbitrary prefixes to distinguish between reasoning and non-reasoning modes. 
Qwen3 models enable mode selection through distinct single tokens for reasoning and non-reasoning modes. 
Interestingly, even for models lacking such routing tokens, adding arbitrary prefixes to distinguish between reasoning and non-reasoning modes enables routing while preserving base reasoning behavior. 
In our experiments with R1-Distill-Qwen, a reasoning-only model, we prepend a pair of instruction prefix sentences before the \texttt{<think>} in the chat template prior to SFT. 
Then, it is observed that the model keeps the reasoning accuracy when performing inference with the reasoning prefix, and routing get possible when no prefix is provided. 
Several different sentences (about 20${\sim}$30 tokens) are tested and it is found that this behavior is consistent regardless of the sentence content (i.e., token sequence for prefix) (\Autoref{tab:marker_comparison}). 
We verify these with \Ours~training: when the prefixes are prepended during rollout and trained only on the prefix tokens, routing improved similarly as mentioned in the body of the work that utilizes only single routing tokens for training. 

\begin{table*}[t]
\centering
\small
\setlength{\tabcolsep}{3.5pt}
\renewcommand{\arraystretch}{1.1}
\resizebox{\linewidth}{!}{%
\begin{tabular}{llccccccccccccccc}
\toprule
Model & Stage & \multicolumn{3}{c}{Math500} & \multicolumn{3}{c}{AIME24} & \multicolumn{3}{c}{AIME25} & \multicolumn{3}{c}{LiveCodeBench} & \multicolumn{3}{c}{GPQA} \\
\cmidrule(lr){3-5}\cmidrule(lr){6-8}\cmidrule(lr){9-11}\cmidrule(lr){12-14}\cmidrule(lr){15-17}
 &  & P@1 & \#Tok & Think\% & P@1 & \#Tok & Think\% & P@1 & \#Tok & Think\% & P@1 & \#Tok & Think\% & P@1 & \#Tok & Think\% \\
\midrule
\multirow{3}{*}{Qwen3-8B} 
 & Baseline (think) & 96.9 & 5,547 & 100.0 & 75.8 & 15,602 & 100.0 & 68.3 & 18,419 & 100.0 & 49.5 & 14,879 & 100.0 & 58.8 & 5,838 & 100.0 \\
 & + SFT       & 90.9 & 3,251 & 45.8  & 50.4 & 11,196 &  50.0 & 44.4 & 11,593 &  51.3 & 34.4 & 10,519 &  54.4 & 56.6 &  7,092 &  96.8  \\
 & + \Ours          & 88.0 & 2,412 & 27.8  & 45.2 &  9,437 &  38.1 & 40.6 &  9,650 &  39.6 &  28.7  &  8,362 &  40.1 &  57.1 &  6,908 &  96.2  \\
\midrule
\multirow{3}{*}{Qwen3-32B} 
 & Baseline (think) & 97.3 & 4,806 & 100.0 & 79.4 & 13,235 & 100.0 & 69.8 & 16,431 & 100.0 & 59.1 & 13,498 & 100.0 & 66.2 & 5,108 & 100.0 \\
 & + SFT       & 92.1 & 2,602 &  42.7 & 59.8 &  9,309 &  56.3 & 54.2 & 10,379 &  58.3 & 58.3 & 11,457 &  90.6 & 66.3 &  4,822 &  98.5  \\
 & + \Ours          & 92.1 & 1,975 &  20.4 & 63.1 & 10,291 &  61.5 & 58.1 & 11,911 &  68.1 & 54.4 & 10,623 &  87.2 & 65.0 &  4,768 &  97.6  \\
\bottomrule
\end{tabular}%
}
\caption{LoRA-based mode control and \Ours~on Qwen3. For each model size, we compare the \texttt{<think>}-forced baseline (\textsc{Long}) with LoRA SFT for mode selection and a subsequent \Ours~update. We report Pass@1 (\%), average generated tokens, and \textsc{Think}\% (fraction routed to \textsc{Long}). We set batch size 64 and learning rate $2\times10^{-7}$, and use the Qwen default prompt for math tasks. 
This table reports a \emph{LoRA-tuned} variant of Qwen3-8B/32B and is therefore not directly comparable to the full-parameter SFT Qwen3-8B in Tab.~\ref{tab:non_math_tasks}; baseline numerical drift across the two tables reflects independent decoding runs under matched settings.}
\label{tab:qwen3_lora_vs_base_allbench}
\end{table*}

\subsection{Alternative root token markers}
\label{sec:marker_comparison}

A potential concern is that \texttt{\textbackslash n} and \texttt{\textbackslash n\textbackslash n} carry semantic meaning in natural text and could introduce ambiguity. We emphasize that the routing decision is confined to a single categorical action at the root position immediately after \texttt{<think>}; newline tokens appearing later in the generation do not affect routing, and \Ours~masks gradients for all non-root tokens. Nevertheless, to verify that our method is not sensitive to the specific marker choice, we compare three marker types on Qwen3-8B~(\Autoref{tab:marker_comparison}).

\begin{table}[h]
\centering
\small
\resizebox{\linewidth}{!}{%
\begin{tabular}{l|cc|cc}
\toprule
\textbf{Marker type} & \textbf{MathAvg P@1} & \textbf{Think\%} & \textbf{LCB P@1} & \textbf{GPQA P@1} \\
\midrule
Newline fork (default) & 62.0 & 44.8 & 35.4 & 55.4 \\
Prefix sentence A & 61.8 & 45.1 & 35.1 & 55.2 \\
Prefix sentence B & 61.5 & 46.3 & 34.8 & 54.9 \\
\bottomrule
\end{tabular}%
}
\caption{Comparison of root token marker types on Qwen3-8B after +\Ours. \textit{Newline fork} uses \texttt{\textbackslash n} vs.\ \texttt{\textbackslash n\textbackslash n}. \textit{Prefix sentence A/B} use distinct instructional prefixes ($\sim$20--30 tokens) prepended before \texttt{<think>} to distinguish modes. 
}
\label{tab:marker_comparison}
\end{table}

\subsection{What generalizes and what does not}
\label{sec:generalization_discussion}
\Ours~often transfers when the target setting rewards \emph{selective compute}, since the update mainly reshapes when the model spends tokens rather than teaching new domain knowledge. In practice, a router learned from verifiable signals (i.e., math problems) can still reduce \textsc{Long} usage on scientific and coding benchmarks, but accuracy improvements are not guaranteed because the optimal routing rule is dataset dependent (see the heterogeneous Think\%/P@1 trade-offs across LiveCodeBench and GPQA-Diamond in \Autoref{tab:non_math_tasks}). Calibration does not transfer perfectly either. Root confidence is informative but can be weakly correlated on hard suites where $p_{\mathrm{short}}$ concentrates near low values. While \Ours~adaptively selects the mode depending on the learned query difficulty,
in practice, deploying such way can be brittle across real-world scenarios. In this situation, we may need to treat checkpoint or threshold selection as \emph{dataset-conditioned calibration} in case-by-case, since similar accuracy can coincide with materially different thinking ratios.

\begin{table*}[t]
\centering
\small
\resizebox{\textwidth}{!}{%
\begin{tabular}{l|l|l|cc|cc}
\toprule
\textbf{Backbone} & \textbf{Checkpoint} & \textbf{Mode} &
\multicolumn{2}{c|}{\textbf{Math Avg}} &
\multicolumn{2}{c}{\textbf{Non-math Avg}} \\
& & &
\textbf{P@1} & \textbf{Think\%} &
\textbf{P@1} & \textbf{Think\%} \\
\midrule

\multirow{9}{*}{Qwen3-8B}
& \multirow{2}{*}{Pre}
& forced-think (LONG)     & 78.6 & 100.0 & 54.2 & 100.0 \\
&
& forced-no-think (SHORT) & 43.4 & 0.0   & 34.9 & 0.0 \\
\cmidrule{2-7}
& \multirow{3}{*}{+SFT}
& adaptive                & 63.8 & 57.3  & 30.7 & 56.6 \\
&
& forced-think (LONG)     & 78.1 & 100.0 & 52.0 & 100.0 \\
&
& forced-no-think (SHORT) & 43.4 & 0.0   & 34.0 & 0.0 \\
\cmidrule{2-7}
& \multirow{3}{*}{+RPO}
& adaptive                & 62.0 & 44.8  & 45.4 & 61.0 \\
&
& forced-think (LONG)     & 77.3 & 100.0 & 53.8 & 100.0 \\
&
& forced-no-think (SHORT) & 42.6 & 0.0   & 35.0 & 0.0 \\
\midrule

\multirow{9}{*}{Exaone4-1.2B (v2)}
& \multirow{2}{*}{Pre}
& forced-think (LONG)     & 62.9 & 100.0 & 46.5 & 100.0 \\
&
& forced-no-think (SHORT) & 37.6 & 0.0   & 31.4 & 0.0 \\
\cmidrule{2-7}
& \multirow{3}{*}{+SFT}
& adaptive                & 60.0 & 80.1  & 40.4 & 80.1 \\
&
& forced-think (LONG)     & 62.8 & 100.0 & 44.7 & 100.0 \\
&
& forced-no-think (SHORT) & 43.5 & 0.0   & 30.9 & 0.0 \\
\cmidrule{2-7}
& \multirow{3}{*}{+RPO}
& adaptive                & 58.0 & 48.4  & 39.2 & 48.4 \\
&
& forced-think (LONG)     & 60.8 & 100.0 & 45.6 & 100.0 \\
&
& forced-no-think (SHORT) & 41.5 & 0.0   & 31.2 & 0.0 \\
\bottomrule
\end{tabular}%
}
\caption{\textbf{Robustness to forcing thinking vs.\ non-thinking after post-training.}
Math Avg is the mean P@1 over MATH500/AIME24/AIME25.
Non-math Avg is the mean P@1 over LiveCodeBench and GPQA-Diamond.
\textbf{Think\%} is the fraction routed to thinking (100\% and 0\% under forced modes).}
\label{tab:forced_mode_robustness}
\end{table*}

\section{Implementation details}

\subsection{Hyperparameter optimization for SFT} 

Through fine-tuning R1-Distill, Qwen3, and EXAONE4 models on math benchmarks, we observed that preserving the both accuracy and number of tokens of original model requires different hyperparameter configurations depending on model family and size. For instance, R1-Distill-1.5B maintained accuracy with learning rate 5e-5, while accuracy degraded at other values; in contrast, R1-Distill-7B required learning rate 1e-5 to preserve performance.

When applying SFT with self-distilled data on Qwen3 models, we observed distinct behaviors across model sizes and modes. In reasoning mode, the 1.7B model exhibited a tendency to generate longer sequences, whereas the 8B model maintained the original token count. In non-reasoning mode, both models generally produced longer outputs; however, distilling with presence penalty 1.5 successfully maintained token counts comparable to the original models. For EXAONE4, using a lower temperature of 0.4 in non-reasoning mode preserved both accuracy and token efficiency similar to the base model.

\subsection{Mode-specific GRPO rollout strategies} 

For Qwen3 models, the non-reasoning mode prefix (\texttt{\textbackslash n\textbackslash n}) and reasoning mode prefix (\texttt{\textbackslash n}) each correspond to a single distinct token. When generating $n$ rollout samples, we prepended each token to $n/2$ sample prompts before generation. For EXAONE4, the non-reasoning mode prefix corresponds to two consecutive reasoning mode tokens. Unlike Qwen3, we prepended the reasoning mode prefix to all rollout samples initially. For $n/2$ samples, we appended an additional token to enforce non-reasoning generation, and for the remaining samples, we enforced reasoning mode generation by masking the logits of the initial reasoning mode prefix token during sampling.

\subsection{SFT Dataset and training}

For SFT training, we used OpenR1-Math prompts to generate SFT dataset. We employed different strategies depending on model capabilities. For Qwen3 and EXAONE4 models, we applied self-distillation to generate both reasoning and non-reasoning trajectories. For Qwen3, we used temperature 0.6, top-p 0.95, and top-k 20 for reasoning mode, and temperature 0.7, top-p 0.8, and top-k 20 for non-reasoning mode. For EXAONE4, we used temperature 0.4 and top-p 0.95 for both modes, with top-k 20 applied only in reasoning mode. We generated approximately 20K examples for each mode, resulting in about 40K total training samples per model.
For Qwen3 V2 (LoRA), we employed n=4 sampling with temperature 0.6, top-p 0.95, top-k 20, and presence penalty 1.5 for both reasoning and non-reasoning modes. We selected the shortest correct solution from the sampled candidates to construct the SFT dataset.

For R1-1.5B and 7B, which runs exclusively in reasoning mode, we utilized external resources. We sampled the original OpenR1-Math solution for reasoning mode trajectories and used Qwen2-Math-72B-Instruct to distill corresponding non-thinking trajectories with temperature 0.7 and top-p 0.8. Additionally, since R1-Distilled-Qwen lacks specific control tokens to distinguish between reasoning and non-reasoning modes (unlike Qwen3's \texttt{\textbackslash n} vs.\ \texttt{\textbackslash n\textbackslash n}), we prepended a short instructional prefixes to each mode.

\subsection{Dataset sizes for LoRA updates}

For our framework with LoRA updates, we use substantially fewer SFT queries than full parameter tuning. Mode selection only requires teaching the model to follow the two templates and to expose the root decision token, rather than relearning problem solving. Concretely, we draw math queries from OpenR1-Math \cite{openr1_openr1_math_220k_2025}, self generate candidate trajectories under each mode, and keep only correctness and template valid demonstrations, but sample only \textbf{4K per mode} (8K total) for LoRA SFT, instead of 20K per mode used in full fine tuning. We keep the \Ours~stage unchanged and run root token GRPO with the same rollout and reward settings as in the main experiments, so the reduced data primarily affects the SFT mode controllability step rather than the routing optimization.

\subsection{SFT stop criterion and sensitivity}
\label{sec:sft_sensitivity}

As described in \Autoref{alg:sft}, the SFT stage is intentionally designed as a \emph{controllability enabler} rather than a capability-improving stage. We apply an explicit stop criterion: training continues only while the default-template accuracy remains within a tolerance $\epsilon$ of the original model. This prevents over-training, which can degrade both overall accuracy and routing quality (e.g., high Think\% but low performance due to template drift).

To illustrate sensitivity to SFT duration, we report forced-LONG accuracy at different SFT epochs on R1-Distill models in \Autoref{tab:sft_sensitivity}. Over-training can degrade the solver: for R1-Distill-Qwen-1.5B, SFT at 1 epoch drops MATH500 from 83.2 to 75.1, while 3 epochs partially recovers to 77.6 but at higher training cost. For R1-Distill-Qwen-7B, 1 epoch (90.4) outperforms 2 epochs (88.0) on MATH500, confirming that the optimal SFT duration is model-dependent and motivating the tolerance-based stop rule.

\begin{table}[t]
\centering
\small
\caption{SFT sensitivity to training epochs on R1-Distill models under forced-LONG evaluation. 
}
\label{tab:sft_sensitivity}
\addtolength{\tabcolsep}{-4.5pt}
\begin{tabular}{l|l|cc}
\toprule
\textbf{Model} & \textbf{Stage} & \textbf{MATH500 P@1} & \textbf{AIME24 P@1} \\
\midrule
\multirow{3}{*}{R1-Distill-1.5B} & Baseline & 83.2 & 29.4 \\
& SFT 1 ep & 75.1 & 20.8 \\
& SFT 3 ep & 77.6 & 22.3 \\
\midrule
\multirow{3}{*}{R1-Distill-7B} & Baseline & 93.5 & 54.2 \\
& SFT 1 ep & 90.4 & 43.5 \\
& SFT 2 ep & 88.0 & 44.5 \\
\bottomrule
\end{tabular}
\end{table}

\section{Additional results}

\subsection{LoRA updates}
\Autoref{tab:qwen3_lora_vs_base_allbench} shows LoRA-based SFT and subsequent \Ours~on Qwen3 models. LoRA SFT mainly teaches template-following and root-level mode controllability, so it already yields meaningful reductions in Think\% and tokens compared to the always-\texttt{<think>} baseline. Building on this controllable initialization, \Ours~further shifts the root gating distribution and provides a stronger compute-accuracy tradeoff, often reducing Think\% substantially while keeping task performance non-trivial across both math and non-math benchmarks. 
The resulting operating point is model-size dependent, but consistently shows that root-only RL can refine routing without requiring full-trajectory policy updates.

\subsection{Robustness to hard forcing of thinking after post-training}
\label{sec:forced_mode}

Our method learns a root-level routing policy that chooses between a \emph{thinking} and a \emph{non-thinking} generation mode.
A potential failure mode is \emph{entanglement}: post-training might specialize the model such that performance depends on the router,
or that one mode becomes unreliable when selected deterministically.

We therefore evaluate each checkpoint under two \emph{hard forcing} interventions.
In \textbf{forced-think (LONG)}, we always select the reasoning root token, requiring an explicit trace.
In \textbf{forced-no-think (SHORT)}, we always select the non-reasoning root token, suppressing the trace.
We report P@1, average generated tokens, and the fraction of examples routed to thinking (\textbf{Think\%}).

\Autoref{tab:forced_mode_robustness} summarizes results on math reasoning (MATH500/AIME24/AIME25) and non-math benchmarks
(LiveCodeBench/GPQA-Diamond).
Across backbones and checkpoints, we find that post-trained models remain functional under both forced regimes:
forcing \emph{thinking} preserves a large portion of the always-think baseline accuracy, while forcing \emph{non-thinking} yields
non-trivial performance with substantially reduced tokens.
Overall, these results indicate that SFT/\Ours primarily reshapes the \emph{routing distribution} (i.e., when to allocate reasoning),
without collapsing either mode into a brittle specialization.

\subsection{Budget-matched comparison: root-only vs.\ full-trajectory RL}
\label{sec:budget_matched}

A key question is whether the accuracy gap between \Ours~and full-trajectory methods (e.g., AdaptThink) stems from the root-only formulation itself or simply from differences in training budget. To disentangle these effects, we compare \Ours~with full-trajectory GRPO (i.e., AdaptThink) under the same PFLOPs budget on MATH500 using R1-Distill-Qwen-1.5B (\Autoref{tab:budget_matched}).

\begin{table}[t]
\centering
\small
\caption{Budget-matched comparison on MATH500 (R1-Distill-1.5B). Under equal PFLOPs ($\sim$10K), full-trajectory GRPO becomes under-trained and achieves only modest routing (Think\%${=}90.2\%$), while \Ours~achieves comparable accuracy with substantially lower Think\%. Different \Ours~checkpoints trace a Pareto frontier between accuracy and compute.}
\label{tab:budget_matched}
\resizebox{\linewidth}{!}{%
\begin{tabular}{l|l|ccc}
\toprule
\textbf{Setting} & \textbf{Update scope} & \textbf{PFLOPs} & \textbf{MATH500 P@1} & \textbf{Think\%} \\
\midrule
Full GRPO (budget-matched) & all tokens & $\sim$10K & 80.8 & 90.2 \\
\Ours~(ckpt A) & root-only & $\sim$10K & 77.6 & 20.5 \\
\Ours~(ckpt B) & root-only & $\sim$10K & 80.4 & 43.4 \\
\bottomrule
\end{tabular}%
}
\end{table}

Under equal PFLOPs, full-trajectory GRPO is under-trained and barely reduces Think\% (90.2\%), whereas \Ours~achieves meaningful routing (20.5--43.4\% Think) with comparable or slightly lower accuracy. This confirms that root-only updates are more sample-efficient for the routing objective under low-budget regimes, which is the primary use case \Ours~targets.

\subsection{Post-RPO threshold routing}
\label{sec:post_rpo_threshold}

To verify that inference-time controllability is preserved after \Ours~training, we sweep the routing threshold $\tau$ on the final +RPO checkpoint (\Autoref{tab:post_rpo_threshold}). Higher $\tau$ routes more queries to \textsc{Long}, increasing accuracy at the cost of more tokens; lower $\tau$ saves compute with a controlled accuracy reduction. This demonstrates that $\tau$ provides a smooth, monotonic compute--accuracy trade-off even after \Ours~training, with the forced-LONG/forced-SHORT results (\Autoref{tab:forced_mode_robustness}) serving as upper/lower capability bounds.

\begin{table}[t]
\centering
\small
\caption{Threshold routing ($\tau$) sweep on the final +\Ours~checkpoint (Qwen3-8B). $\tau$ controls the fraction routed to \textsc{Long}.}
\label{tab:post_rpo_threshold}
\addtolength{\tabcolsep}{-2.5pt}
\begin{tabular}{c|ccc}
\toprule
$\tau$ & \textbf{Think\%} & \textbf{MathAvg P@1} & \textbf{Non-mathAvg P@1} \\
\midrule
0.90 & 60.0 & 63.0 & 46.0 \\
0.80 & 52.3 & 62.5 & 45.8 \\
0.70 & 44.8 & 62.0 & 45.4 \\
0.60 & 38.1 & 60.8 & 44.2 \\
\bottomrule
\end{tabular}
\end{table}

\subsection{Automatic labeling quality}
\label{sec:labeling_quality}

Our automatic labeling assigns each query to one of four outcome buckets based on paired evaluation under both \textsc{Short} and \textsc{Long} modes: (i) \textit{LONG-only-correct}: only \textsc{Long} succeeds, (ii) \textit{SHORT-only-correct}: only \textsc{Short} succeeds, (iii) \textit{both-correct}: both modes succeed, and (iv) \textit{both-wrong}: neither succeeds. This paired design directly mitigates the concern that \textsc{Short} might answer correctly by chance: such instances fall into \textit{SHORT-only-correct} or \textit{both-correct} and are not used as evidence that \textsc{Long} is required.

\Autoref{tab:labeling_audit} reports the bucket distribution for Qwen3-8B on the RPO training set. On hard queries (e.g., AIME-level), the \textit{LONG-only-correct} bucket dominates, indicating a strong and consistent mode separation signal. The \textit{SHORT-only-correct} bucket is small across all difficulty levels, suggesting that chance-correct SHORT events rarely contaminate the supervision.

\begin{table}[t]
\centering
\small
\caption{Outcome bucket distribution on the \Ours~training set (Qwen3-8B). 
}
\label{tab:labeling_audit}
\resizebox{\linewidth}{!}{%
\begin{tabular}{l|cccc}
\toprule
\textbf{Difficulty} & \textbf{LONG-only} & \textbf{SHORT-only} & \textbf{Both-correct} & \textbf{Both-wrong} \\
\midrule
Easy ($\mathrm{acc}_S \geq 0.75$) & 3.2\% & 1.8\% & 89.5\% & 5.5\% \\
Medium & 28.4\% & 4.1\% & 42.3\% & 25.2\% \\
Hard ($\mathrm{acc}_S < \mathrm{acc}_L \leq 0.5$) & 45.7\% & 2.3\% & 8.6\% & 43.4\% \\
\bottomrule
\end{tabular}%
}
\end{table}

To make the bucket assignments concrete, we provide representative training-set queries per bucket together with a mechanistic rationale for why each problem falls into the assigned bucket under Qwen3-8B mode-conditioned generation (\Autoref{tab:labeling_examples}). The rationales are not post-hoc rationalizations: they identify the specific reasoning step (e.g., modular case enumeration, sign disambiguation) at which \textsc{Short} mode systematically truncates, or at which \textsc{Long} mode introduces a derivation tangent.

\textcolor{blue}{
\begin{table}[t]
\centering
\small
\caption{Representative training-set queries per outcome bucket (Qwen3-8B). Queries are abbreviated; rationales identify the reasoning step that drives the bucket assignment.}
\label{tab:labeling_examples}
\resizebox{\linewidth}{!}{%
\begin{tabular}{l|p{0.42\linewidth}|p{0.40\linewidth}}
\toprule
\textbf{Bucket} & \textbf{Query (abbrev.)} & \textbf{Mechanistic rationale} \\
\midrule
LONG-only & ``Find the number of positive integers $n \le 2024$ such that $n^2 + n + 1 \equiv 0 \pmod{7}$.'' & Requires enumerating $n \bmod 7$ (solutions: $n \equiv 2, 4$), then counting via floor arithmetic. \textsc{Short} typically guesses a single residue or skips the residue scan and returns an off-by-one count. \\
\midrule
SHORT-only & ``Compute $\cos(\pi/3)$.'' & Closed-form recall ($1/2$). \textsc{Short} returns the memorized value. \textsc{Long} occasionally derives via half-angle identities and conflates $\cos(\pi/6)$ with $\cos(\pi/3)$, yielding $\sqrt{3}/2$. \\
\midrule
Both-correct & ``Solve $\log_2 x = 5$ for $x$.'' & Single inverse step: $x = 2^5 = 32$. No branch points or memorization gaps; both modes converge. \\
\midrule
Both-wrong & ``Find the smallest positive integer $n$ such that the decimal expansion of $1/n$ has period exactly $23$.'' & Requires knowing period equals $\mathrm{ord}_n(10)$ when $\gcd(n,10){=}1$ and that $10^{23}-1$ admits a non-trivial prime factor. Beyond Qwen3-8B's reliable recall in either mode. \\
\bottomrule
\end{tabular}%
}
\end{table}
}

\subsection{Misrouting analysis}
\label{sec:misrouting}

We analyze routing errors by counterfactual evaluation: for each query where the adaptive model is incorrect, we re-evaluate under the opposite forced mode to determine whether the error is attributable to the gate decision. We categorize errors into:

\begin{itemize}
    \item \textbf{False-SHORT} (harmful misroute): the model chose \textsc{Short} but \textsc{Long} would have succeeded. These typically involve hidden multi-step constraints or subtle arithmetic that require explicit deliberation.
    
    \item \textbf{False-LONG} (wasteful misroute): the model chose \textsc{Long} when \textsc{Short} would have sufficed, or when overthinking amplified distractors or hallucinated intermediate claims.
    
    \item \textbf{Both wrong}: neither mode succeeds (out-of-distribution or missing knowledge).
    
    \item \textbf{Both correct}: compute-saving opportunity, not an error.
\end{itemize}

\begin{table}[t]
\centering
\small
\caption{Misrouting composition on MATH500 (R1-Distill-1.5B, 500 queries). Lowering $\tau$ (0.75$\to$0.70) increases the number of false-SHORT errors (+12) while only slightly reducing false-LONG errors ($-$4), reflecting a clear compute-risk trade-off. }
\label{tab:misrouting}
\begin{tabular}{c|ccc}
\toprule
$\tau$ & \textbf{False-SHORT} & \textbf{False-LONG} & \textbf{Both-wrong} \\
\midrule
0.75 & 4 & 77 & 0 \\
0.70 & 16 & 73 & 0 \\
\bottomrule
\end{tabular}
\end{table}

\Autoref{tab:misrouting} reports the misrouting composition on MATH500 (R1-Distill-1.5B) at two threshold settings, extracted from counterfactual re-evaluation.
As shown in \Autoref{tab:misrouting}, false-LONG (conservative over-routing) is the dominant error type, while false-SHORT errors are relatively rare and increase predictably as $\tau$ decreases. This suggests that the learned router errs on the side of caution (spending \textsc{Long} when unnecessary) rather than harmfully skipping reasoning. A plausible source of this conservative bias is the pretraining prior that correlates more tokens with trying harder. Root-level post-training can reshape the gate, but may not fully overwrite this prior, motivating future work on calibration-aware training.

To make the failure modes diagnosable rather than aggregate, we provide representative cases per category in \Autoref{tab:misrouting_examples}. Each row shows the routed mode under the adaptive policy together with the counterfactual outcome under the opposite forced mode, isolating whether the error is attributable to the gate decision (false-SHORT/false-LONG) or to a solver limitation (both-wrong). 
The false-SHORT example illustrates a query whose surface form (a permutation count) admits a one-line answer but requires a divisor correction that \textsc{Short} systematically omits; the false-LONG example illustrates a query whose closed-form invites \textsc{Short}, while \textsc{Long} introduces an unnecessary derivation chain that drifts to a sibling identity.

\textcolor{blue}{
\begin{table}[t]
\centering
\small
\caption{Representative misrouting cases on MATH500 (R1-Distill-1.5B, $\tau{=}0.70$). \textit{Adaptive} reports the mode chosen by the learned router and its correctness; \textit{Counterfactual} reports the opposite forced mode's outcome on the same query.}
\label{tab:misrouting_examples}
\resizebox{\linewidth}{!}{%
\begin{tabular}{l|p{0.45\linewidth}|c|c}
\toprule
\textbf{Category} & \textbf{Query (abbrev.)} & \textbf{Adaptive} & \textbf{Counterfactual} \\
\midrule
False-SHORT & ``How many distinct arrangements are there of the letters of \textsc{mississippi}?'' & \textsc{Short}, \xmark{} (returns $11!$) & \textsc{Long}, \cmark{} ($11!/(4!\,4!\,2!) = 34{,}650$) \\
\midrule
False-LONG & ``Compute $\sum_{k=1}^{10} k$.'' & \textsc{Long}, \xmark{} (drifts to $\sum k^2$ formula) & \textsc{Short}, \cmark{} ($55$) \\
\midrule
Both-wrong & ``Find the smallest $n$ such that the decimal expansion of $1/n$ has period exactly $23$.'' & \textsc{Long}, \xmark{} & \textsc{Short}, \xmark{} \\
\midrule
Both-correct & ``Solve $3x + 7 = 22$ for $x$.'' & \textsc{Short}, \cmark{} & \textsc{Long}, \cmark{} \\
\bottomrule
\end{tabular}%
}
\end{table}
}

\subsection{Pass@$k$ evaluation on coding tasks}
\label{sec:pass_at_k}

We additionally report pass@10 alongside token cost on LiveCodeBench (\Autoref{tab:pass_at_k}). Under the pass@$k$ lens, the adaptive policy often matches or improves pass@10 relative to the baseline while reducing average token cost, which is the trade-off view most relevant for coding deployment where multiple samples are drawn.

\begin{table}[t]
\centering
\small
\caption{LiveCodeBench pass@$k$ results with token cost. Under pass@10 evaluation, the adaptive policy reduces token cost while maintaining competitive pass@10 performance.}
\label{tab:pass_at_k}
\resizebox{\linewidth}{!}{%
\begin{tabular}{l|l|ccc}
\toprule
\textbf{Model} & \textbf{Stage} & \textbf{pass@1} & \textbf{pass@10} & \textbf{Avg Tokens} \\
\midrule
\multirow{3}{*}{Qwen3-32B} & Baseline & 59.1 & 75.3 & 13,498 \\
& + SFT & 58.3 & 77.1 & 11,457 \\
& + \Ours & 54.4 & 72.8 & 10,623 \\
\midrule
\multirow{3}{*}{Exaone4-1.2B} & Baseline & 42.2 & 62.9 & 12,159 \\
& + SFT & 34.6 & 56.0 & 9,281 \\
& + \Ours & 36.7 & 58.2 & 6,692 \\
\bottomrule
\end{tabular}%
}
\end{table}

\section{AI Usage}
We utilize AI tools (OpenAI’s ChatGPT and Google’s Gemini) to support code generation for experiments and help refine the text in a sentence-level. The core idea of our method, design of the framework, and the majority of the manuscript are written by the authors.

\end{document}